# The Metric-FF Planning System: Translating "Ignoring Delete Lists" to Numeric State Variables

**Jörg Hoffmann**                                                    HOFFMANN@INFORMATIK.UNI-FREIBURG.DE
*Institut für Informatik*
*Georges-Köhler-Allee, Geb. 52*
*79110 Freiburg*
*Germany*

## Abstract

Planning with numeric state variables has been a challenge for many years, and was a part of the 3rd International Planning Competition (IPC-3). Currently one of the most popular and successful algorithmic techniques in STRIPS planning is to guide search by a heuristic function, where the heuristic is based on relaxing the planning task by ignoring the delete lists of the available actions.

We present a natural extension of "ignoring delete lists" to numeric state variables, preserving the relevant theoretical properties of the STRIPS relaxation under the condition that the numeric task at hand is "monotonic". We then identify a subset of the numeric IPC-3 competition language, "linear tasks", where monotonicity can be achieved by pre-processing. Based on that, we extend the algorithms used in the heuristic planning system FF to linear tasks. The resulting system Metric-FF is, according to the IPC-3 results which we discuss, one of the two currently most efficient numeric planners.

## 1. Introduction

The planning community has long been aware of the fact that purely propositional representation languages, in particular STRIPS (Fikes & Nilsson, 1971), are not well suited for modeling various phenomena that are essential in real-world problems. In particular, modeling context dependent effects, concurrent execution of actions with different duration, and continuous resources are all awkward, or impossible, within the STRIPS language. To overcome the first of these limitations, Pednault (1989) defined the (nowadays widely accepted) ADL language, which amongst other things allows for conditional effects (effects that only occur when their condition holds true in the state of execution). To overcome (one or both of) the latter two limitations, various proposals have been made (e.g., Ghallab & Laruelle, 1994; Koehler, 1998; Smith & Weld, 1999). The most recent effort in this direction is the PDDL2.1 language defined by Fox and Long (2002) as the input language for the 3rd International Planning Competition (IPC-3). The IPC series is a biennial challenge for the planning community, inviting planning systems to participate in a large scale publicly accessible evaluation. IPC-3 was hosted at AIPS-2002, and stressed planning beyond the STRIPS formalism, featuring tracks for temporal and numeric planners. This article describes the approach behind one of the planners that participated in IPC-3, Metric-FF. Metric-FF is an extension of the FF system (that can handle ADL) to numeric constructs.

Currently one of the most popular and successful algorithmic techniques in STRIPS planning is to guide search (forward or backward, state space or plan space) by a heuristic function, where the heuristic is based on relaxing the planning task by ignoring the delete lists (i.e. the negative effects) of the available actions. The heuristic value of a search state in this framework is (an estimate of) the difficulty of extending the state to a solution using the relaxed actions. This idea was first, independently, proposed by McDermott (1996) and Bonet et al (1997), and is now widely used in a huge number of variations. Examples of planners that use the idea are Unpop (McDermott, 1996, 1999), HSP in its various configurations (Bonet & Geffner, 1998, 1999, 2001), GRT (Refanidis & Vlahavas, 1999, 2001), MIPS (Edelkamp & Helmert, 2001), STAN4 (Fox & Long, 2001), RePOP (Nguyen & Kambhampati, 2001), Sapa (Do & Kambhampati, 2001), and FF (Hoffmann, 2000; Hoffmann & Nebel, 2001). The search paradigms used by these planners include forward and backward state space search as well as partial-order planning. The forward state space planner FF was especially successful at IPC-2 (Bacchus, 2001). In what follows we extend the heuristic idea for STRIPS, ignoring delete lists, to numeric state variables in a way that preserves the relevant theoretical properties of the STRIPS relaxation. We phrase these properties *admissibility*, *basic informedness*, and *polynomiality*. While the investigation takes place in the setting of forward state space search as used by FF, it seems likely that the same ideas will also work in other search schemes such as plan-space search (some more on this in the outlook, Section 8). The Sapa system also deals with numeric constructs. The heuristic function, however, completely ignores numeric goals and thus lacks one of the relevant theoretical properties, basic informedness (we will return to this later). There are also numeric versions of MIPS and GRT. On the respective MIPS version there is no publication available at the time of writing but an article (Edelkamp, 2003), which the reader is referred to, is to appear in this same JAIR special issue. The numeric version of GRT, GRT-R (Refanidis & Vlahavas, 2000), allows only for a restricted form of numeric variables and expressions, basically a limited form of resource allocation and consumption. The heuristic function considers resource consumption as another form of state cost. This, like Sapa's heuristic, lacks basic informedness, as we will see later.

In a numeric planning task, there can be numeric constraints (in action preconditions and the goal) and numeric effects (in action effects). Constraints and effects can be of different types. For example, a constraint can require that the value of a variable be either at least as high as or at most as high as a given constant. The numeric effects can, from a semantic perspective, either increase or decrease the value of the affected variable. Now, the delete effects in STRIPS decrease the logical value of the propositional variables, so the idea we explore is to relax the numeric task by ignoring all decreasing effects. The main difficulty with this idea is that ignoring the decreasing effects does not necessarily simplify the task. For example, when the goal requires that $x < 0$ and $x$ is initially equal to 0, the decreasing effects are needed to solve the task, so the relaxed task is unsolvable. The relaxation is thus only adequate (preserves the theoretical properties mentioned above) in tasks where it is always preferable to have higher variable values. We call such tasks *monotonic*.[1] We observe that tasks that belong to a subset of the numeric IPC-3 competition language, *linear tasks* (in which the numeric variables are only used in linear functions), can be brought into a normal form that is monotonic. Based on that, we extend the heuristic algorithms used in FF, and thereby the whole system, to linear tasks.

---

1. There is a duality here with respect to ignoring the *in*creasing effects or the *de*creasing effects. If lower variable values are always preferable then ignoring the increasing effects is an adequate relaxation. Whether one chooses one or the other does not seem to make much difference. We choose monotonicity in the positive sense only because it is conceptually simpler.

FF (Hoffmann & Nebel, 2001) is a close relative of HSP (Bonet & Geffner, 2001). Search takes place forward in the state space, i.e., starting from the initial state new states are explored until a goal state is found. The search process in FF is guided by a heuristic function that is based on solving, in each search state $s$, the relaxed task starting from $s$. The heuristic value to $s$ is the number of actions in the respective relaxed plan, i.e., the number of actions needed to achieve the goal from $s$ when assuming the delete lists are all empty. States with lower heuristic value are preferred. The main obstacle in the extension of FF to numeric state variables is to extend the machinery that solves the relaxed task in each search state. Once this machinery is defined, the rest of the system translates effortlessly. We evaluate the resulting planning system Metric-FF by discussing the results of the numeric domains used in the 3rd International Planning Competition. As it turns out, Metric-FF and LPG (Gerevini, Saetti, & Serina, 2003a) were the best performing numeric planners in the competition.[2]

The article is structured as follows. Throughout the text we refer to related work where it is relevant. We first give the necessary background in terms of STRIPS notation, and the techniques that the STRIPS version of FF uses. Section 3 introduces our notation for numeric state variables, i.e., for the numeric subset of PDDL2.1. Section 4 describes how the heuristic principle for STRIPS, the relaxation, can be extended to the numeric setting. Section 5 defines our algorithms for solving relaxed numeric tasks. Section 6 then fills in the details on how the relaxed plans are used to implement the Metric-FF planning system, and we briefly describe how ADL constructs can be handled, and how flexible optimization criteria can be taken into account. The IPC-3 results are discussed in Section 7. Section 8 concludes and outlines future work. An appendix contains most proofs.

## 2. STRIPS Techniques

In this section, we give background on the techniques that the FF system uses in the STRIPS language. We start by examining the relaxation that underlies FF's heuristic function. We then proceed to the algorithms that are used to solve relaxed tasks. We finally describe how the relaxed plans are used to implement the actual FF system. The discussion is a little more detailed than would strictly be necessary to understand the FF workings. This serves to provide a solid background for what is to come: Sections 4, 5, and 6 will, in turn for each of the subtopics dealt with in this section, show how these methodologies can be extended to the numeric setting.

Before we start, we give the notation for the STRIPS language. When we refer to sets we mean finite sets. We consider the propositional STRIPS language, where all constructs are based on logical propositions. A world *state* $s$ is a set of (the true) propositions. An *action* $a$ is given as a triple of proposition sets, $a = (pre(a), eff(a)^+, eff(a)^-)$: $a$'s *precondition, add list*, and *delete list*, respectively (we use the somewhat unusual notation $eff(a)^+$ and $eff(a)^-$ as this makes the extension to numeric variables more readable).

---

We first specify the semantics of world states and actions. Throughout the article, we consider sequential planning only, where a single action at a time is applied to a world state.[3] Actions induce state transitions as follows. Given a world state $s$ and an action $a$, the *result* of executing (the action sequence consisting solely of) $a$ in $s$, $result(s, \langle a \rangle)$, is $result(s, \langle a \rangle) := s \setminus eff(a)^- \cup eff(a)^+$ if the action is *applicable* in $s$, $pre(a) \subseteq s$. Otherwise, $result(s, \langle a \rangle)$ is undefined. The result of executing an action sequence $\langle a_1, \ldots, a_m \rangle$ in a state is recursively defined by $result(s, \langle a_1, \ldots, a_m \rangle) := result(result(s, \langle a_1, \ldots, a_{m-1} \rangle), a_m)$, and $result(s, \langle \rangle) = s$.

A STRIPS task – we use the word "task" rather than "problem" to avoid confusion with the complexity theoretic notion of decision problems – is a tuple $(P, A, I, G)$: the set $P$ of logical propositions used in the task, the set $A$ of actions, the *initial state $I$* (a world state), as well as the *goal $G$* (a partial world state, see below). All propositions in the actions, initial state, and goal are taken from $P$. Given a task $(P, A, I, G)$, what one wants to find is a *plan*. An action sequence $\langle a_1, \ldots, a_m \rangle \in A^*$ is a plan for $(P, A, I, G)$ if $G \subseteq result(I, \langle a_1, \ldots, a_m \rangle)$. Since the $\subseteq$ relation (not equality) is used here, there could be several *goal states* in which a plan ends. If there exists at least one plan for a task, then the task is *solvable*. Sometimes we refer to *optimal* plans. In our sequential framework, a plan is optimal for a task if there is no plan for the task that contains fewer actions.

## 2.1 Relaxing Strips Tasks

We want to inform the search for a plan by a function that estimates the goal distance of search states. The idea is to define a relaxation (i.e., a simplification) of planning tasks, then solve, in any search state, the relaxed task, and take the length of the relaxed solution as an estimate of how long the solution from the state at hand really is. The relaxation that was first proposed by McDermott (1996) and Bonet, Loerincs, & Geffner (1997), is to relax STRIPS tasks by ignoring the delete lists of all actions.

**Definition 1** *Assume a STRIPS task $(P, A, I, G)$. The* relaxation $a^+$ *of an action $a \in A$, $a = (pre(a), eff(a)^+, eff(a)^-)$, is defined as*

$$a^+ := (pre(a), eff(a)^+, \emptyset).$$

*The* relaxation *of $(P, A, I, G)$ is $(P, A^+, I, G)$, where $A^+ := \{a^+ \mid a \in A\}$. An action sequence $\langle a_1, \ldots, a_n \rangle \in A^*$ is a* relaxed plan *for $(P, A, I, G)$ if $\langle a_1^+, \ldots, a_n^+ \rangle$ is a plan for $(P, A^+, I, G)$.*

Ignoring the delete lists simplifies the task because the action preconditions and the goal are all positive. We identify a number of desirable properties that the relaxation has. We will later define relaxations for numeric variables that have the same properties.

**Definition 2** *Let RPLANSAT denote the following problem.*

Assume a STRIPS task $(P, A, I, G)$. Is the relaxation of $(P, A, I, G)$ solvable?

---

3. As opposed to, e.g., Graphplan-based approaches (Blum & Furst, 1997), which find sets of actions to be applied in parallel.

**Proposition 1** *The relaxation given in Definition 1 is adequate, i.e., the following holds true.*

1. *Admissibility: any plan that solves the original task also solves the relaxed task, i.e., assuming a STRIPS task $(P, A, I, G)$, any plan for $(P, A, I, G)$ is also a relaxed plan for $(P, A, I, G)$.*

2. *Basic informedness: the preconditions and goals can trivially be achieved in the original task if and only if the same holds in the relaxed task, i.e., assuming a STRIPS task $(P, A, I, G)$, $\langle \rangle$ is a plan for $(P, A, I, G)$ if and only if $\langle \rangle$ is a relaxed plan for $(P, A, I, G)$, and for $a \in A$, $result(I, \langle \rangle) \supseteq \text{pre}(a)$ if and only if $result(I, \langle \rangle) \supseteq \text{pre}(a^+)$.*

3. *Polynomiality: the relaxed task can be solved in polynomial time, i.e., deciding RPLANSAT is in P.*

The proof is trivial – admissibility and basic informedness follow directly from the definitions, and polynomiality was proved earlier by Bylander (1994). The proof can be found in Appendix A.

If we want to use the length of relaxed plans as a heuristic function, the properties stated by Proposition 1 are important for the following reasons. Admissibility tells us that optimal relaxed plan length is an admissible heuristic, since the optimal real plan is also a relaxed plan.[4] Also, we will not mistake a solvable state for a dead end: if there is no relaxed plan then there is no real plan either (more on this below). The "only if" directions in basic informedness tell us that the relaxation does not give us any constraints for free (for example, the heuristic value will be zero only in goal states). If the heuristic does not have these properties then possibly parts of the problem must be solved in regions where there is no heuristic information at all (like when the heuristic value is already zero but no goal state is reached yet).[5] Polynomiality tells us that we can compute the heuristic function efficiently.

## 2.2 Solving Relaxed Tasks

Ideally, given a search state $s$, we would like to know how many relaxed actions are *at least* needed to reach the goal, i.e., we would like to know what the length of an *optimal* relaxed plan is (this would be an admissible heuristic, c.f. above). But finding optimal relaxed plans is still intractable (Bylander, 1994). So instead we compute arbitrary, i.e., not necessarily optimal, relaxed plans. This is done with a Graphplan-style algorithm (Blum & Furst, 1997; Hoffmann & Nebel, 2001). Given a search state $s$ in a STRIPS task $(P, A, I, G)$, we first build a relaxed planning graph starting from $s$, i.e., for the task $(P, A, s, G)$. Then we extract a relaxed plan from that graph. The graph building algorithm is depicted in Figure 1.

---

4. Note that using the term "admissibility" this way slightly abuses notation, as admissibility usually refers to a property of the heuristic function, not the technique (relaxation, in our case) it is based on.

5. The formulation of basic informedness might seem unnecessarily complicated. We chose the general formulation at hand so that the definition can easily be transferred to other relaxation techniques, like the ones we introduce later.

$P_0 := s$
$t := 0$
**while** $G \nsubseteq P_t$ **do**
    $A_t := \{a \in A \mid pre(a) \subseteq P_t\}$
    $P_{t+1} := P_t \cup \bigcup_{a \in A_t} eff(a)^+$
    **if** $P_{t+1} = P_t$ **then** fail **endif**
    $t := t + 1$
**endwhile**
$finallayer := t$, succeed

Figure 1: Building a relaxed planning graph for a task $(P, A, s, G)$.

The planning graph in the relaxed case is simply represented as a sequence $P_0$, $A_0$, ..., $A_{t-1}$, $P_t$ of proposition sets and action sets. These are built incrementally in the obvious fashion, starting with $P_0 = s$ as the initial layer, and iteratively inserting the add effects of all applicable actions. The algorithm fails if at some point before reaching the goals no new propositions come in. This only happens when the relaxed task is unsolvable.

**Proposition 2** *Assume a STRIPS task $(P, A, I, G)$, and a state $s$. If the algorithm depicted in Figure 1 fails, then there is no relaxed plan for $(P, A, s, G)$.*

The proof is in Appendix A. The main argument is that, if two consecutive proposition layers are identical, then the same will hold true at all later layers so the graph has reached a fixpoint.

In case the goals can be reached at layer $finallayer$, we call the relaxed plan extraction mechanism depicted in Figure 2. The *level* of each proposition $p$ (action $a$) here is the first layer in the relaxed planning graph at which $p$ ($a$) appears, i.e., the minimum $t$ such that $p \in P_t$ ($a \in A_t$).

**for** $t := 1, \ldots, finallayer$ **do**
    $G_t := \{g \in G \mid level(g) = t\}$
**endfor**
**for** $t := finallayer, \ldots, 1$ **do**
    **for** all $g \in G_t$ **do**
        select $a$, $level(a) = t - 1$, $g \in eff(a)^+$
        **for** all $p \in pre(a)$ **do**
            $G_{level(p)} \cup = \{p\}$
        **endfor**
    **endfor**
**endfor**

Figure 2: Extracting a relaxed plan for a task $(P, A, s, G)$ (levels and $finallayer$ computed by the algorithm in Figure 1).

Relaxed plan extraction is based on a sequence $G_1, \ldots, G_{finallayer}$ of goal and sub-goal sets. Goals and sub-goals are always inserted into the set at their respective level, i.e., at the position of their first appearance in the relaxed planning graph. The goal sets are initialized by inserting the respective (top-level) goals. A backwards loop then selects, at each layer, actions to support the respective goal set. All goals or sub-goals $g$ are supported, and the preconditions of the respective actions become new sub-goals. This way, upon termination the selected actions can be used to form a relaxed plan for the state at hand.

**Proposition 3** *Assume a STRIPS task $(P, A, I, G)$, and a state $s$ for which the algorithm depicted in Figure 1 reaches the goals. The actions selected by the algorithm depicted in Figure 2 form a relaxed plan for $(P, A, s, G)$.*

As all goals and sub-goals are supported, arranging the actions selected at each layer in an arbitrary order yields a relaxed plan. The proof is in Appendix A.

## 2.3 FF

Based on the relaxed plan information, a heuristic state space planner is easily implemented. Choices must be made on how to use the relaxed plans, and how to arrange the search strategy. We describe the specific methods used in FF, which are very efficient in many STRIPS and ADL benchmarks (Hoffmann & Nebel, 2001). The extended system uses straightforward adaptions of these methods. We define a heuristic function, a search strategy, and a pruning technique. The heuristic estimates goal distances as relaxed plan length.

**Definition 3** *Assume a STRIPS task $(P, A, I, G)$, and a state $s$. The FF heuristic value $h(s)$ for $s$ is defined as follows. If the algorithm depicted in Figure 1 fails, $h(s) := \infty$. Otherwise, $h(s) := \sum_{t=1}^{finallayer} |A_t|$ where $A_t$ is the set of actions selected at layer $t$ by the algorithm depicted in Figure 2.*

If there is no relaxed plan for a state, then the heuristic value is set to $\infty$. This is justified by the first property proved in Proposition 1: when there is no relaxed plan then there can be no real plan either, i.e., the state is a *dead end* in the sense that the goals can not be reached from it. Such states can be pruned from the search. The search scheme we use is a kind of hill-climbing procedure using a complete lookahead to find better states. See Figure 3.

Enforced hill-climbing, like (standard) hill-climbing, starts out in the initial state and performs a number of search iterations trying to improve on the heuristic value, until a state with zero value is reached. While normally, iterative improvement is done by selecting one best direct successor of the current search state, enforced hill-climbing uses a complete breadth first search to find a strictly better, possibly indirect, successor. The search cuts out states that have been seen earlier during the same iteration, and does not expand states that the heuristic function recognizes as dead ends. This strategy works well when the better successors are usually nearby, which is the case in many planning benchmarks when using the FF heuristic function (Hoffmann, 2001, 2002b). When there is no better successor below the current search node, the algorithm fails (more on this below).

We finally define a pruning technique, selecting a set of the most promising successors to each search state. The unpromising successors can then be ignored. A promising successor is a state generated by an action that is helpful in the following sense.

```
initialize the current plan to the empty plan  <>
s := I
while h(s) ≠ 0 do
        starting from s, perform breadth first search for a state s' with h(s') < h(s),
                        avoiding repeated states using a hash table,
                        not expanding states s'' where h(s'') = ∞
        if no such state can be found then fail endif
        add the actions on the path to s' at the end of the current plan
        s := s'
endwhile
output current plan, succeed
```

Figure 3: The enforced hill-climbing algorithm, for a task with heuristic $h$.

**Definition 4** *Assume a STRIPS task $(P, A, I, G)$, and a state $s$ for which the algorithm depicted in Figure 1 reaches the goals. The set of helpful actions $H(s)$ to $s$ is defined as*

$$H(s) := \{a \in A \mid \text{eff}^+(a) \cap G_1 \neq \emptyset\},$$

*where $G_1$ is the set of sub-goals constructed at layer 1 by the algorithm depicted in Figure 2.*

In other words, an action is considered helpful if it achieves at least one of the lowest level goals in the relaxed plan to the state at hand. The helpful actions information is used as a pruning technique. During a breadth first search iteration in enforced hill-climbing, when expanding a state $s$, only the states generated by actions in $H(s)$ are included into the search space. Note that states $s$ where the relaxed planning graph does not reach the goals have $h(s) = \infty$ so do not get expanded anyway.

In general, neither enforced hill-climbing nor helpful actions pruning maintain completeness. The algorithm fails if enforced hill-climbing gets caught in a dead end state. This can happen because the search does not backtrack over its decisions, and because the heuristic function can return a value below $\infty$ for a dead end state. The algorithm also fails if helpful actions pruning cuts out important states, which can happen because the technique is a non-admissible approximation of usefulness. We deal with this issue by employing a safety-net solution, i.e., if enforced hill-climbing fails then the planner starts from scratch using a complete heuristic search engine, without any pruning technique. The search engine used is what Russel and Norvig (1995) term *greedy best-first* search. This is a weighted $A^*$ strategy where the weight $w_g$ of the node cost in the equation $f(s) = w_g * g(s) + w_h * h(s)$ is $w_g = 0$, i.e., search simply expands all search nodes by increasing order of goal distance estimation. Repeated states are avoided in the obvious way by keeping a hash table of visited states.

## 3. Numeric State Variables

We introduce notation for the numeric part of the PDDL2.1 language (i.e., PDDL2.1 level 2) defined by Fox and Long (2002), used at IPC-3. We restrict ourselves to STRIPS for readability reasons. Extensions to ADL will be summarized in Section 6.2. All sets are assumed to be finite unless stated otherwise.

In addition to the propositions $P$, we now have a set $V$ of numeric variables. Notationally, we say $V = \{v^1, \ldots, v^n\}$ (throughout the article, $n$ will denote the number of numeric variables). A *state* $s$ is a pair $s = (p(s), v(s))$ where $p(s) \subseteq P$ is a set of propositions and $v(s) = (v^1(s), \ldots, v^n(s)) \in \mathbf{Q}^n$ is a vector of rational numbers (the obvious semantics being that $p(s)$ are the true propositions, and $v^i(s)$ is the value of $v^i$).[6]

An *expression* is an arithmetic expression over $V$ and the rational numbers, using the operators $+$, $-$, $*$, and $/$. A *numeric constraint* is a triple $(exp, comp, exp')$ where $exp$ and $exp'$ are expressions, and $comp \in \{<, \leq, =, \geq, >\}$ is a *comparator*. A *numeric effect* is a triple $(v^i, ass, exp)$ where $v^i \in V$ is a variable, $ass \in \{:=, +=, -=, *=, /=\}$ is an *assignment operator*, and $exp$ is an expression (the effect right hand side). A *condition* is a pair $(p(con), v(con))$ where $p(con) \subseteq P$ is a set of propositions and $v(con)$ is a set of numeric constraints. An *effect* is a triple $(p(eff)^+, p(eff)^-, v(eff))$ where $p(eff)^+ \subseteq P$ and $p(eff)^- \subseteq P$ are sets of propositions (the add- and delete-list), and $v(eff)$ is a set of numeric effects such that $i \neq j$ for all $(v^i, ass, exp), (v^j, ass', exp') \in v(eff)$.[7] An *action* $a$ is a pair $(pre(a), eff(a))$ where $pre(a)$ is a condition and $eff(a)$ is an effect.

The semantics of this language is straightforward. The *value* $exp(v)$ of an expression $exp$ in a variable value vector $v$ (in $s$, if $v$ is the numeric part $v(s)$ of a state $s$) is the rational number that the expression simplifies to when replacing all variables with their respective values, or undefined if a division by 0 occurs. A constraint $(exp, comp, exp')$ *holds* in a state $s$, written $s \models (exp, comp, exp')$, if the values of $exp$ and $exp'$ are defined in $s$, and stand in relation $comp$ to each other. A condition $con = (p(con), v(con))$ holds in a state $s$, $s \models con$, if $p(con) \subseteq p(s)$, and all numeric constraints in $v(con)$ hold in $s$. The value $(v^i, ass, exp)(v)$ of a numeric effect $(v^i, ass, exp)$ in a variable value vector $v$ (in $s$, if $v$ is the numeric part $v(s)$ of a state $s$) is the outcome of modifying the value of $v^i$ in $s$ with the value of $exp$ in $s$, using the assignment operator $ass$. A numeric effect is *applicable* in $s$ if its value in $s$ is defined. An effect $eff = (p(eff)^+, p(eff)^-, v(eff))$ is applicable in $s$ if all numeric effects in $v(eff)$ are applicable in $s$. For such $eff$ and $s$, $eff(s)$ is the state $s'$ where $p(s') = p(s) \setminus p(eff)^- \cup p(eff)^+$, and $v(s')$ is the value vector that results from $v(s)$ when replacing $v^i(s)$ with $(v^i, ass, exp)(s)$ for all $(v^i, ass, exp) \in v(eff)$. Putting all of these definitions together, the *result* of executing an action $a$ in a state $s$ is $result(s, \langle a \rangle) = eff(a)(s)$ if $s \models pre(a)$ and $eff(a)$ is applicable in $s$, undefined otherwise. In the first case, $a$ is said to be applicable in $s$. For an action sequence $\langle a_1, \ldots, a_n \rangle$, $result(s, \langle a_1, \ldots, a_n \rangle)$ is as usual defined recursively by $result(s, \langle a_1, \ldots, a_n \rangle) = result(result(s, \langle a_1, \ldots, a_{n-1} \rangle), a_n)$ and $result(s, \langle \rangle) = s$.

A *numeric task* is a tuple $(V, P, A, I, G)$ where $V$ and $P$ are the variables and propositions used, $A$ is a set of actions, $I$ is a state, and $G$ is a condition. A sequence of actions $\langle a_1, \ldots, a_n \rangle \in A^*$ is a *plan* if the result of applying it to $I$ yields a state that models $G$, $result(I, \langle a_1, \ldots, a_n \rangle) \models G$.

In our algorithmic framework, we make distinctions between different degrees of expressivity that we allow in numeric constraints and effects, i.e., between different *numeric*

---

6. We ignore, for readability reasons, the possibility given in Fox and Long's original language that a variable can have an undefined value until it is assigned one. Our methodology can be easily extended – and is in fact implemented – to deal with this case.

7. Fox and Long (2002) make this assumption implicitly, by requiring that the outcome of an action is well-defined – note that commutative effects on the same variable can be merged.

*languages.* A numeric language is a tuple (*Cons*, *Eff-ass*, *Eff-rh*) where *Cons* is a possibly infinite set of numeric constraints, *Eff-ass* is a set of assignment operators, and *Eff-rh* is a possibly infinite set of expressions. A task $(V, P, A, I, G)$ belongs to a language if all constraints, assignment operators, and effect right hand sides are members of the respective sets.

The next three sections contain the technical part of this article. They are organized as follows.

1. Section 4 provides the theory on which Metric-FF's heuristic function is based. The relaxation, ignoring delete lists as described in Section 2.1, is extended to numeric variables. Section 4.1 formalizes the key idea in a restricted numeric language, and states that the extended relaxation fulfills admissibility, basic informedness, and polynomiality. Section 4.2 abstracts from the restricted language, identifying generalized semantic properties that make the relaxation work. Section 4.3 then introduces the language of linear tasks, which can be brought into a linear normal form (LNF) that has these semantic properties. Metric-FF's core planning algorithms are implemented for LNF tasks.

2. Section 5 introduces the algorithms implemented in Metric-FF's heuristic function. The algorithms are extensions to the relaxed Graphplan methods described in Section 2.2. Section 5.1 describes the algorithms for the restricted language, Section 5.2 extends that to LNF tasks. We state formally that the algorithms are complete and correct. We also see that the algorithms are, in theory, less efficient than they could be. The number of relaxed planning graph layers built can be exponential in the size of the task encoding, in contrast to polynomiality of the relaxation as proved in Section 4. The reason why the implementation lags behind what is theoretically possible is that the implementation work was done before the theory was fully developed. However, from a practical point of view, it is at least debatable how important the potential exponentiality is (the number of relaxed planning graph layers built is bounded by the length of a shortest relaxed plan). Exploring this issue in depth is a topic for future work. More details are in Sections 4.1 and 5.1.

3. Section 6 details how the relaxed plan information is used to implement the Metric-FF system. Section 6.1 explains the extension of the basic FF architecture as described in Section 2.3. Section 6.2 explains the extension to ADL, Section 6.3 describes how flexible optimization criteria can be dealt with.

## 4. Relaxing Numeric State Variables

We show how the relaxation technique for STRIPS can naturally be extended to the numeric context. We proceed in the three steps outlined above.

### 4.1 A Restricted Language

The key idea in our relaxation becomes apparent when one considers the context where constraints only compare variables to constants via $\geq$ or $>$, there are only += and -=

effects, and the effect right hand sides are positive constants. More formally, our restricted language is:

( $\{(v^i, comp, c) \mid v^i$ variable, $comp \in \{\geq, >\}, c \in \mathbf{Q}\}$,
  $\{+=, -=\}$,
  $\{c \mid c \in \mathbf{Q}, c > 0\}$ )

In STRIPS, delete lists are troublesome because they falsify propositions that we might need for preconditions or the goal. In the restricted numeric language here, the -= effects are troublesome because they diminish the value of the affected variables. The idea is therefore to ignore these effects.

**Definition 5** *Assume a restricted numeric task* $(V, P, A, I, G)$. *The* relaxation $a^+$ *of an action* $a \in A$, $a = (\text{pre}(a), (p(\text{eff}(a))^+, p(\text{eff}(a))^-, v(\text{eff}(a))))$, *is defined as*

$$a^+ := (\text{pre}(a), (p(\text{eff}(a))^+, \emptyset, \{(v^i, +=, \exp) \mid (v^i, +=, \exp) \in v(\text{eff}(a))\})).$$

*The relaxation of* $(V, P, A, I, G)$ *is* $(V, P, A^+, I, G)$, *where* $A^+ := \{a^+ \mid a \in A\}$. *An action sequence* $\langle a_1, \ldots, a_n \rangle \in A^*$ *is a* relaxed plan *for* $(V, P, A, I, G)$ *if* $\langle a_1^+, \ldots, a_n^+ \rangle$ *is a plan for* $(V, P, A^+, I, G)$.

The above relaxation is adequate in the restricted language, in the precise sense introduced in Section 2.1.

**Definition 6** *Let RESTRICTED-RPLANSAT denote the following problem.*

Assume a restricted numeric task $(V, P, A, I, G)$. Is the relaxation of $(V, P, A, I, G)$ solvable?

**Theorem 1** *The relaxation given in Definition 5 is adequate, i.e., the following holds true.*

1. Admissibility: *assuming a restricted numeric task* $(V, P, A, I, G)$, *any plan for* $(V, P, A, I, G)$ *is also a relaxed plan for* $(V, P, A, I, G)$.

2. Basic informedness: *assuming a restricted numeric task* $(V, P, A, I, G)$, $\langle \rangle$ *is a plan for* $(V, P, A, I, G)$ *if and only if* $\langle \rangle$ *is a relaxed plan for* $(V, P, A, I, G)$, *and for* $a \in A$, $result(I, \langle \rangle) \models \text{pre}(a)$ *if and only if* $result(I, \langle \rangle) \models \text{pre}(a^+)$.

3. Polynomiality: *deciding RESTRICTED-RPLANSAT is in P.*

The detailed proof can be found in Appendix A. It is a straightforward extension of the STRIPS proof, exploiting the correspondence between pre/goal-conditions, add lists, and delete lists on the one hand, and $x \geq [>]c$ constraints, += effects, and -= effects on the other hand. The only tricky part lies in proving polynomiality, precisely in how to handle repeated increasing effects on the same variable. Such effects might have to be applied an exponential number of times. Consider the tasks, for $n \in \mathbf{N}_0$, where $v^i$ is initially 0, $v^i \geq n$ is the goal, and we have an action effect $(v^i, +=, 1)$. For task $n$, the shortest relaxed plan comprises $n$ steps, which is exponentially long in the size of a non-unary encoding

301

of $n$. The trick one can use to decide relaxed solvability in polynomial time is a simple $\infty$ handling. The polynomial decision process is a forward fixpoint procedure similar to building a relaxed planning graph. As soon as there appears an action $a$ that increases a variable $v^i$, one can assume that $v^i$'s value is $\infty$, reflecting the fact that $v^i$'s value can be made arbitrarily high by applying $a$ a sufficient number of times. As indicated earlier, the current implementation of Metric-FF, which we will describe in Section 5, does *not* make use of such an $\infty$ handling technique, and may thus build an exponential number of relaxed planning graph layers for a search state. More on this in Section 5.1.

A few words on related work are in order here. If one relaxes numeric tasks by ignoring all the numeric constructs, then one gets admissibility and polynomiality, but not basic informedness. The heuristic methods used in Sapa (Do & Kambhampati, 2001) and GRT-R (Refanidis & Vlahavas, 2000) come quite close to this extreme case. In fact, Sapa's heuristic constructs a relaxed plan that completely ignores the numeric part of the task. Then the "resource consumption" of the resulting relaxed plan (roughly, the sum of all decreasing effects on numeric variables) is used to estimate the number of actions that would be needed to re-produce these resources, and that number is added to the heuristic value of the state at hand. In particular, this method ignores all numeric goals and preconditions and thus lacks basic informedness. Similarly, the heuristic technique used in GRT-R considers resource consumption as another form of state cost, but does not take any numeric precondition or goal constraints into account. The heuristic technique does not make explicit use of relaxed plans so our definitions can not be directly applied. However, as numeric constraints are not considered, the heuristic value of a purely numeric action precondition is zero even if the precondition is not true in the current state, and the technique thus also lacks basic informedness.

## 4.2 Monotonicity, and a Dynamic Relaxation

We now have a look behind the scenes of the relaxation technique that we used above for the restricted language. We abstract from the syntax of the numeric constructs, and focus on their semantics instead. We define an extension of our relaxation to the general context, and identify a group of semantic properties that make this relaxation adequate. We will later focus on a syntactically restricted language, linear tasks, where it is easier to see that the relaxation is adequate. The main intention of the abstract work in this subsection is to provide some theoretical background on the general characteristics for which our relaxation works.

Let us first ignore semantic issues, and simply extend the definition of our relaxation. In general, the definition is not as easy as for the restricted case in Definition 5. While our idea is still to ignore decreasing effects, the difficulty is that whether an effect is decreasing or not can depend on the context it is executed in.[8] As a simple example, say an action $a$ has a numeric effect $(v^i, +=, v^j)$. If $v^j$ has a negative value in the state of $a$'s execution, this effect decreases the value of $v^i$ instead of increasing it. So we can not statically relax $a$ by ignoring parts of its specification. Instead, our relaxation now is dynamic: we relax the state transition function.

---

8. It is common practice to refer to $+=$ effects as "increasing effects", and to $-=$ effects as "decreasing effects". In contrast to that, we distinguish between syntax and semantics by using $+= / -=$ to denote syntax, and increasing / decreasing to denote semantics (of arbitrary numeric effects).

**Definition 7** *Assume a state $s$ and an action $a = (\text{pre}(a), \text{eff}(a))$. The relaxed result of executing $a$ in $s$ is $result^+(s, a) = s'$ such that $p(s') = p(s) \cup p(\text{eff}(a))^+$, and $v(s')$ is the value vector that results from $v(s)$ when replacing $v^i(s)$ with $(v^i, \text{ass}, \exp)(s)$ for all $(v^i, \text{ass}, \exp) \in v(\text{eff})$ such that $(v^i, \text{ass}, \exp)(s) > v^i(s)$.*

For an action sequence $\langle a_1, \ldots, a_n \rangle$, $result^+(s, \langle a_1, \ldots, a_n \rangle)$ is defined recursively as with the original $result$ function in Section 2. Note that, in STRIPS and the restricted numeric language, Definition 7 comes down to exactly the relaxations we have used before.

Having generalized our relaxation, we now want to know in exactly which situations this relaxation is adequate. Obviously, ignoring the decreasing effects is not adequate in general. As a simple example, if the value of a variable $v^i$ is initially 0, there is an effect $(v^i, \text{-=}, 1)$, and the goal requires that $v^i < 0$, then the "relaxation" renders the task unsolvable. Intuitively, the relaxation is adequate if it is always preferable for the numeric variables to have higher values. Formalizing this intuition turns out to be a bit tricky. Recall our three conditions for adequacy of a relaxation: admissibility (any real plan is also a relaxed plan), basic informedness (the relaxation does not ignore any precondition or goal constraints), and polynomiality (solvability of the relaxation can be decided in polynomial time). Basic informedness is obviously given for our relaxation here. Not so admissibility and polynomiality. Say we want to make sure that each real plan is also a relaxed plan. Not only must the numeric constraints prefer higher variable values, but the effects must also. As an example, say we have $v^i = v^j = 0$ initially, the goal $v^i \geq 1$, an action effect $(v^i, \text{-=}, v^j)$, and an action effect $(v^j, \text{-=}, 1)$. If we ignore the decreasing effect on $v^j$, we can not solve the task because for the effect $(v^i, \text{-=}, v^j)$ it is better when $v^j$ takes on lower values. Considering polynomiality, to ensure that relaxed solvability can be decided in polynomial time, all kinds of subtleties must be handled. Say we want to shortcut repeated action application by an $\infty$ trick, i.e., by assuming that repeated application of increasing effects makes the affected variable diverge (as is the case in the restricted language above). Then we will get in trouble if repeated (relaxed) application of an action makes the value of the affected variable converge.[9] Similar difficulties arise when an expression in a constraint does not diverge with its variables. Finally, it might be that the constraint looks correct when inserting $\infty$, but can never be fulfilled with finite values. An example of this is the constraint $v^i \geq v^i + 1$, which is fulfilled when inserting $\infty$ for $v^i$.

In the following definition, we introduce a number of conditions that are sufficient to ensure that none of the difficulties described above appear. We will see that in "monotonic" tasks each real plan is also a relaxed plan, and that in "strictly monotonic" tasks, given their := effects are acyclic in a certain sense, relaxed plan existence can be decided in polynomial time.

**Definition 8** *Assume a numeric task $(V, P, A, I, G)$. The task is* monotonic *if, for all pairs of states $s$ and $s'$ with $\forall v^i : v^i(s) \leq v^i(s')$, the following holds.*

---

9. As an example, if $v^i$ is initially 1 and we have an effect $(v^i, \text{+=}, 1 - \frac{v^i}{2})$ then repeated application of the effect makes the value of $v^i$ converge to 2 (the value of $v^i$ after $n$ applications is $2 - \frac{1}{2}^n$).

*(1)* *For all numeric constraints* $(\exp, \text{comp}, \exp')$ *occurring in the task:*

$$s \models (\exp, \text{comp}, \exp') \Rightarrow s' \models (\exp, \text{comp}, \exp').$$

*(2)* *For all numeric effects* $(v^i, \text{ass}, \exp)$ *occurring in the task:*

$$(v^i, \text{ass}, \exp)(s) \leq (v^i, \text{ass}, \exp)(s'),$$

 *where the* $\leq$ *relation holds only if both values are defined.*

*The task is* strongly monotonic *if the above and the following hold.*

*(3)* *For all states* $s$ *and* $s'$ *as above, for all numeric effects* $(v^i, \text{ass}, \exp)$ *occurring in the task, with* ass $\in \{+=, -=, *=, /=\}$:

$$(v^i, \text{ass}, \exp)(s) - v^i(s) \leq (v^i, \text{ass}, \exp)(s') - v^i(s'),$$

 *where the* $\leq$ *relation holds only if both values are defined.*

*(4)* *For all expressions* $\exp$ *occurring in the task:*

$$\forall v^i \in v(\exp) : \lim_{v^i \to \infty} \exp = \infty,$$

 *where* $v(\exp)$ *denotes the set of all variables contained in* $\exp$.

*(5)* *For all numeric constraints* $(\exp, \text{comp}, \exp')$ *occurring in the task:*

$$\exists s : s \models (\exp, \text{comp}, \exp').$$

Some explanation of this lengthy definition is in order. Condition (1) ensures that the numeric constraints prefer higher variable values. Condition (2) does the same for effects, requiring that the value of an effect can only increase with the variables. In particular, the value does not become undefined, i.e., no division by zero occurs when the variables grow. These two conditions suffice to make each real plan a relaxed plan, as higher variable values are always preferable. Conditions (3) to (5) aim at making relaxed solvability easy to decide. Condition (3) is a stronger version of condition (2). We require that the *value that the effect adds to the affected variable* increases with the variables. This ensures that repeated application of the effect causes the value of the affected variable to diverge. To illustrate this, an effect $(v^i, +=, -v^j + c)$ fulfills condition (2) but not condition (3). The outcome of this effect is always $c$, which is monotonic in its (zero) variables but affects $v^i$ more as $v^i$'s own value becomes higher. Condition (4) postulates that all expressions diverge in all variables, and condition (5) postulates that to all constraints there is a finite variable assignment that makes these constraints true. Together with condition (1) these requirements ensure that the constraints will eventually be fulfilled when increasing the values of the variables.[10]

---

10. One could weaken conditions (1) to (3) of Definition 8 by exploiting the fact that we are only interested in *reachable* states. It does not matter if, e.g., a constraint is not monotonic in a region of variable values that will never be reached due to the semantics of the task. Metric-FF implements no such analysis techniques, except throwing away actions – and with them, numeric constraints and effects – whose preconditions can not be reached in the relaxed planning graph for the initial state, when ignoring all numeric constructs. Exploring the topic in more depth is future work.

The := effects are separated out from the definition of strong monotonicity, i.e., while we do postulate condition (2) for them, we do not postulate condition (3). Postulating condition (3) for := effects would also suffice. But this condition does not hold for even the simplest form of := effects, namely $(v^i, :=, c)$, assigning a constant to a variable. Note that this is in principle the same effect as the example given above, $(v^i, +=, -v^j + c)$. Effects of this kind are common even in the limited suits of benchmarks that are currently available (e.g., when filling up a tank, the fuel level is assigned the maximum level). So we identify a different sufficient criterion that makes := effects tractable, and that captures the common forms of these effects. Computing the maximum outcome of a set of assignment effects, in the relaxation and under condition (2), becomes easy if the value changes on each single variable can not be propagated into their own value. The proof argument is that, if, transitively, a change on $v^i$ can not influence $v^i$'s own value then all one needs to do is to perform value propagation steps, at each step computing the maximum assignment available for each variable. After at most as many steps as there are variables, the values will be fixed. We formalize the possible value propagations with a straightforward graph definition.

**Definition 9** *Assume a numeric task $(V, P, A, I, G)$. The task has* acyclic := effects *if the graph $(V, E)$ is cycle-free, where*

$$E = \{(v^i, v^j) \in V \times V \mid \exists a \in A, (v^j, :=, \exp) \in v(\mathrm{eff}(a)) : v^i \in v(\exp)\},$$

*with $v(\exp)$ denoting the set of all variables contained in* exp.

We now state in which ways our definitions imply adequacy of ignoring decreasing effects as a relaxation. In the notation for the relaxed plan existence decision problem, we abstract from syntactic issues, and assume that a well-formed input task to the decision procedure is strongly monotonic and has acyclic := effects.

**Definition 10** *Let STRONGLY-MONOTONIC-RPLANSAT denote the following problem.*

Assume a numeric task $(V, P, A, I, G)$. Is there a relaxed plan for $(V, P, A, I, G)$, provided the task is strongly monotonic and has acyclic := effects?

**Theorem 2** *The relaxation given in Definition 7 is adequate for strongly monotonic tasks with acyclic := effects. Precisely the following holds true.*

1. Admissibility: *assuming a monotonic numeric task $(V, P, A, I, G)$, any plan for $(V, P, A, I, G)$ is also a relaxed plan for $(V, P, A, I, G)$.*

2. Basic informedness: *assuming a numeric task $(V, P, A, I, G)$, $\langle\rangle$ is a plan for $(V, P, A, I, G)$ if and only if $\langle\rangle$ is a relaxed plan for $(V, P, A, I, G)$, and for $a \in A$ $result(I, \langle\rangle) \models \mathrm{pre}(a)$ if and only if $result^+(I, \langle\rangle) \models \mathrm{pre}(a)$.*

3. Polynomiality: *deciding STRONGLY-MONOTONIC-RPLANSAT is in P.*

The proof, given in Appendix A, is basically a straightforward exploitation of the properties ensured by the above definitions. Note that Theorem 2 only identifies sufficient criteria that make our relaxation work. Interesting questions are, are there other, maybe weaker, criteria? As a concrete example, there seem to be certain cases of cyclic assignment effects that can be easily handled. What exactly are these cases? Answering these questions is a topic for future work.

Another thing we have not dealt with is how our semantic constraints translate to the syntax of the arithmetic expressions that are allowed in PDDL2.1. We do not consider the details of this but base the rest of the article on a subset of PDDL2.1 where the required semantic properties can easily be achieved – the language for which the Metric-FF system is actually implemented. Extending the system to richer languages is an open research topic.

## 4.3 Linear Tasks, and LNF

The Metric-FF system is implemented to deal with what we call *linear* tasks. This is the language of numeric tasks where there are no $*=$ or $/=$ effects, and the numeric variables are only used in linear expressions. More formally:

$(\ \{(exp, comp, exp') \mid exp, exp' \text{ linear expression}, comp \text{ arbitrary}\},$
$\{:=, +=, -=\},$
$\{exp \mid exp \text{ linear expression}\}\ )$

Metric-FF's implementation allows for tasks that are linear after the following pre-processing step. Assume we are given a planning task $(V, P, A, I, G)$. A variable $v^i \in V$ is a *task constant* if $v^i$ is not affected by the effect of any action in $A$. An expression is a task constant if all variables occurring in it are task constants. The pre-process replaces all task constants with the respective rational numbers resulting from inserting the initial variable values.[11]

Linear tasks are, of course, not necessarily monotonic. In fact, all of the illustrative counter examples we have given above are linear. But linear functions are monotonic, more precisely strictly monotonic and diverging, in all variables, either in the positive or in the negative sense. The idea is to introduce, for a variable $v^i$ that is used in the negative sense at some point, an *inverted* variable $-v^i$ that always takes on the value $(-1) * v^i$. One can then replace $v^i$ with $-v^i$ at the points where $v^i$ is used negatively. When this has been done for all variables, the task is (strictly) monotonic: all variables are only used in the positive sense (more details below). Introducing inverted variables can be viewed as a shortcut way of informing the heuristic function about which places to use the variables in the positive or in the negative sense.[12] We will return to this issue when considering Metric-FF's heuristic algorithms in Section 5.2.

Given a linear task, Metric-FF transforms the task into what we call its *linear normal form* (LNF). In an LNF task, the expressions are weighted sums of variables, where the weights are all greater than 0. The transformation process works as follows. First, a series of simple steps transforms the task into the following language.

---

11. If, in a quotient $(exp/exp')$, $exp'$ simplifies to 0 then the expression is undefined and the respective constraint can never be fulfilled / the respective action's effects can never become applicable. In this case one can replace the constraint with "false"/ remove the action.

12. David Smith, personal communication.

( $\{(\sum_{j \in X} c^j * v^j + c, \geq [>], 0) \mid c^j, c \in \mathbf{Q}, c^j \neq 0\}$,
  $\{:=, +=\}$,
  $\{\sum_{j \in X} c^j * v^j + c \mid c^j, c \in \mathbf{Q}, c^j \neq 0\}$ )

To achieve this language format, one replaces all constraints $(exp, =, exp')$ with $(exp, \leq, exp')$ and $(exp, \geq, exp')$, and all effects $(v^i, \text{-=}, exp)$ with $(v^i, +=, -exp)$. The rest is a matter of normalizing linear functions. The language format differs from LNF only in that the variable weights may be negative. This, of course, makes all the difference. Reconsider the example where a variable $v^i$ is initially 0, there is an action $a$ with effect $(v^i, \text{-=}, 1)$, and the goal requires that $v^i < 0$. We take this as the running example in the following. In the above language format, $a$'s effect is $(v^i, +=, -1)$, and the goal requires that $(-1) * v^i > 0$. Due to the negative weighting of $v^i$ in the goal condition, ignoring decreasing effects is not viable.

The way we introduce inverted variables is an extension to the methodology that eliminates negative preconditions in STRIPS planning (a technique first introduced by Gazen & Knoblock, 1997). The process works as follows. Initialize the set $T$ of translated variables to $T := \emptyset$. Iterate until there are no more negative weights, otherwise select an (arbitrary) occurrence $c^j * v^j$, $c^j < 0$, in a weighted sum. Introduce a new variable $-v^j$. Set $-v^j(I) := (-1) * v^j(I)$. For all effects $(v^j, +=[:=], \sum_{j \in X} c^j * v^j + c)$, introduce (into the effect set of the same action) the effect $(-v^j, +=[:=], \sum_{j \in X}((-1) * c^j) * v^j + ((-1) * c))$. Set $T := T \cup \{v^j, -v^j\}$. For all occurrences of $c * v$ in weighted sums such that $c < 0$ and $v \in T$ (where $v$ may be one of the original variables or one of the introduced inverse variables), replace $c * v$ with $((-1) * c) * -v$ (where $-v$ is the respective inverse counterpart to $v$). After at most $|V|$ iterations, all weights are positive and the process terminates. The task is then in the following linear normal form.

( $\{(\sum_{j \in X} c^j * v^j + c, \geq [>], 0) \mid c^j, c \in \mathbf{Q}, c^j > 0\}$,
  $\{:=, +=\}$,
  $\{\sum_{j \in X} c^j * v^j + c \mid c^j, c \in \mathbf{Q}, c^j > 0\}$ )

For our running example, the LNF transformation is the following. There are now two variables, $v^j$ and $-v^j$, both of which are initially 0. The action $a$ has two effects, namely $(v^i, +=, -1)$ and $(-v^i, +=, 1)$. The goal condition is now expressed in terms of the value of $-v^i$, and reads $-v^i > 0$. A single application of the action achieves the goal, as it also does under the relaxed transition function because the effect on $-v^i$ is increasing.[13] In general, it is easy to see that LNF tasks are strongly monotonic.

**Proposition 4** *Assume a linear numeric task $(V, P, A, I, G)$. If the task is in LNF, then it is strongly monotonic.*

**Proof:** All conditions in Definition 8 are trivially fulfilled in LNF tasks. As examples, condition (1) is true because we only compare expressions that are (positively) monotonic in all variables to constants via $\geq$ or $>$. Condition (3) is true because we only have $+=$ effects whose right hand sides are (positively) monotonic in all variables. □

---

13. Note that estimating the maximum value of $-v^i$ is the same as estimating the minimum value of $v^i$. More on this in Section 5.2.

With Proposition 4, if an LNF task $(V, P, A, I, G)$ has acyclic := effects (remember that these are separated out from Definition 8 condition (3)) then the task fulfills the prerequisites of Theorem 2, so ignoring the decreasing effects is an adequate relaxation. It is thus feasible to use solutions to the relaxation as a means of heuristic estimation.

For the := effects, one can easily translate these into, e.g., += effects – $(v^i, :=, \exp)$ translates to $(v^i, +=, ((-1) * v^i) + \exp)$. So the reader might wonder why we bother treating := effects at all. The point is that, while the translated effects behave equivalently under the real transition function, they behave differently in the relaxation. In our running example suppose there is a second action $a'$ with the effect $(v^i, :=, 10)$. In the LNF transformation, the translated version of this effect is $(v^i, +=, -v^i + 10)$. Say we execute, under $result^+$, first $a$ (with effects $(v^i, +=, -1)$ and $(-v^i, +=, 1)$) then $a'$. In the original task, the resulting value of $v^i$ is 10. In the translated task, that value is 11 (because the decreasing effect on $v^i$ is ignored). So it does make a difference whether we treat := effects separately or not. An open question is whether, or in which situations, that difference is important for planner performance.

We also remark that, while Metric-FF implements the introduction of inverted variables for LNF tasks only, it seems likely that similar processes will work for richer languages, when all functions are strictly monotonic and diverging in all variables.

## 5. Solving Relaxed Tasks

We now concentrate on the algorithms used in FF, more generally algorithms that can be used to obtain heuristic information in a forward state space search. We explain how to solve relaxed numeric tasks. We first consider the restricted language, then extend the methods to LNF tasks. The algorithms form the basis of the Metric-FF implementation.

### 5.1 Restricted Tasks

The implementation uses a straightforward extension of the Graphplan-style algorithms introduced in Section 2.2. We still use a two-step process that first builds a relaxed planning graph then extracts a relaxed plan from that (if the graph succeeds in reaching the goals). In parallel to the structures that keep track of the progress in logical propositions, we now have structures that keep track of the progress in terms of maximally possible variable values. The graph building mechanism is outlined in Figure 4.

The parts of the algorithm concerned with the propositions work exactly as in the STRIPS case, c.f. Section 2.2. As for the numeric variables, the *max* value vector at a layer $t$ specifies the current maximum value that the variables can take on. The vectors are updated in the obvious fashion, adding at each layer the total sum of the increasing effects at that layer. The termination condition now checks whether the maximum values of all variables have either not changed, or are already higher than needed: the mneed$^i$ value for each variable $v^i$ is defined as the highest requirement on that variable, i.e.,

$$\text{mneed}^i := \max(-\infty, \{c \mid (v^i, \geq [>], c) \in v(G) \cup \bigcup_{a \in A} v(pre(a))\}).$$

Note that the algorithm fails only if there is no relaxed plan for $(V, P, A, s, G)$: if the algorithm fails at a layer $m$ then the termination condition will hold true at all later layers.

$P_0 := s$, **for** all $v^i$ **do** $max_0^i := v^i(s)$ **endfor**
$t := 0$
**while** $p(G) \not\subseteq P_t$ or $(v^i, \geq [>], c) \in v(G), max_t^i \not\geq [\not>]c$ **do**
    $A_t := \{a \in A \mid p(pre(a)) \subseteq P_t,$
                    $\forall (v^i, \geq [>], c) \in v(pre(a)) : max_t^i \geq [>]c\}$
    $P_{t+1} := P_t \cup \bigcup_{a \in A_t} p(eff(a))^+$
    **for** all $v^i$ **do** $max_{t+1}^i := max_t^i + \sum_{a \in A_t, (v^i, +=, c) \in v(eff(a))} c$ **endfor**
    **if** $P_{t+1} = P_t$ and
        $\forall v^i : max_{t+1}^i = max_t^i$ or $max_t^i > \text{mneed}^i$ **then**
      **fail**
    **endif**
    $t := t + 1$
**endwhile**
$finallayer := t$

Figure 4: Building a relaxed planning graph for a state $s$ in a restricted numeric task $(V, P, A, I, G)$.

Also, note that there can be only a finite number of layers as the numeric variables that do increase will eventually reach their finite mneed values. But, as mentioned in Section 4.1, the number of layers can be exponential in the task encoding. Reconsider the example where, for $n \in \mathbf{N}_0$, $v^i$ is initially 0, $v^i \geq n$ is the goal, and we have an action effect $(v^i, +=, 1)$. The number of graph layers built for this example, $n$, is exponential in a non-unary encoding of $n$, whereas one could easily decide solvability with the $\infty$ trick outlined in Section 4.1. On the other hand, it appears unlikely that an implementation of the provably polynomial decision procedure would be better in practice. The graph building algorithm is polynomial in the length of its output (the minimal length of a relaxed plan). Also, the possibly exponential minimal length of a relaxed plan (exponential in a non-unary encoding of the variable values) does not seem particularly relevant, at least not in examples that are not specifically constructed to provoke this exponentiality. It remains an open question whether an implementation of $\infty$ handling can achieve better performance in realistic examples.

We now focus on relaxed plan extraction. This is invoked if the relaxed planning graph succeeds in reaching the goals. The information that the graph provides us with are the *levels* of all actions, propositions, and numeric goals. For actions and propositions the level is the first graph layer at which they appear, c.f. Section 2.2. For numeric goals $(v^i, \geq [>], c)$, the level is the graph layer $t$ where the goal can first be achieved, i.e., where $max_t^i \geq [>]c$ holds the first time. The plan extraction mechanism is outlined in Figure 5.

Again, the logical entities are dealt with exactly as in the STRIPS case, c.f. Section 2.2. In addition to the propositional (sub-)goal set $p(G_t)$ at each layer $t$ we now have a set $v(G_t)$ of numeric goals. Like in STRIPS, goals and sub-goals are always inserted into the set at their first appearance in the relaxed planning graph, and the goal sets are initialized by inserting the respective (top-level) goals. Then there is a backwards loop from the top to the bottom layer, selecting actions to support the propositions and numeric variables in the respective goal sets. The propositions are supported as before, the only difference being **for** $t := 1, \ldots, finallayer$ **do**
    $p(G_t) := \{g \in p(G) \mid level(g) = t\}$
    $v(G_t) := \{(v^i, \geq [>], c) \in v(G) \mid level(v^i, \geq [>], c) = t\}$
**endfor**
**for** $t := finallayer, \ldots, 1$ **do**
    **for** all $g \in p(G_t)$ **do**
        select $a$, $level(a) = t - 1$, $g \in p(eff(a))^+$
        **for** all $p \in p(pre(a)), (v^i, \geq [>], c) \in v(pre(a))$ **do**
            $p(G_{level(p)}) \cup = \{p\}$
            $v(G_{level(v^i, \geq [>], c)}) \cup = \{(v^i, \geq [>], c)\}$
        **endfor**
    **endfor**
    **for** all $(v^i, \geq [>], c) \in v(G_t)$ **do**
        **while** $max_{t-1}^i \not\geq [\not>] c$ **do**
            select $a$, $level(a) = t - 1$, $(v^i, +=, c') \in v(eff(a))$,
                  $a$ not previously selected in this while-loop
            $c := c - c'$
            /* introduce $a$'s preconditions as above */
        **endwhile**
        $v(G_{t-1}) \cup = \{(v^i, \geq [>], c)\}$
    **endfor**
**endfor**

Figure 5: Extracting a relaxed plan for a state $s$ in a restricted numeric task $(V, P, A, I, G)$ (levels and $finallayer$ computed by the algorithm in Figure 4).

that now also the numeric preconditions of the supporting actions must be inserted into the goal sets below. When uniting sets of numeric goals that both contain a constraint on the same variable $v^i$, the stronger one of both constraints is taken. For the numeric goals $(v^i, \geq [>], c) \in v(G_t)$ it is in general not enough to select a single action as several actions at $t - 1$ might have contributed to $v^i$'s maximum value at $t$. So supporters are selected until the goal can be achieved one layer earlier. Note that $max_t^i - \sum_{a \in A_{t-1}, (v^i, +=, c) \in v(eff(a))} c = max_{t-1}^i$, so the **while** loop will always terminate successfully. Note also that one occurrence of an action can support different logical and numeric goals by different effects, but can not be used to support the same numeric goal twice.

Upon termination of plan extraction, the selected actions can be used to form a relaxed plan: with $A_t$ denoting the actions selected at layer $t$, an arbitrary linearization of $A_0, \ldots, A_{finallayer-1}$ is a relaxed plan for the task. Note that one can apply various simple heuristics, like selecting $+=$ effects with maximum right hand side first, to make the relaxed plans as short as possible.

## 5.2 LNF Tasks

The algorithms for numeric tasks in linear normal form differ from those for restricted tasks in that we need to take care of $:=$ effects, and of the more general expressions in numeric constraints and in effect right hand sides. As it turns out, integrating these extensions is not overly difficult. The only issue that becomes slightly involved is the exact termination criterion for relaxed graph building. In our solution to the issue we assume, as in the theoretical analysis underlying Theorem 2, that the := effects are acyclic. An outline of the graph building mechanism is shown in Figure 6.

$P_0 := s$, **for** all $v^i$ **do** $max_0^i := v^i(s)$ **endfor**
$t := 0$
**while** $p(G) \not\subseteq P_t$ or $(exp, \geq [>], 0) \in v(G), exp(max_t) \not\geq [\not>]0$ **do**
$\quad A_t := \{a \in A \mid p(pre(a)) \subseteq P_t,$
$\quad\quad\quad\quad\quad \forall (exp, \geq [>], 0) \in v(pre(a)) : exp(max_t) \geq [>]0\}$
$\quad P_{t+1} := P_t \cup \bigcup_{a \in A_t} p(eff(a))^+$
$\quad$**for** all $v^i$ **do** $max_{t+1}^i := max_t^i + \sum_{a \in A_t : (v^i, +=, exp) \in v(eff(a)), exp(max_t) > 0} exp(max_t)$ **endfor**
$\quad$**for** all $v^i$ **do** $max_{t+1}^i := \max(max_{t+1}^i, \max_{a \in A_t, (v^i, :=, exp) \in v(eff(a))} exp(max_t))$ **endfor**
$\quad$**if** $P_{t+1} = P_t$ and
$\quad\quad \forall v^i : max_{t+1}^i = max_t^i$ or $max_t^i > \text{mneed}^i(s)$ **then**
$\quad\quad$**fail**
$\quad$**endif**
$\quad t := t + 1$
**endwhile**
$finallayer := t$

Figure 6: Building a relaxed planning graph for a state $s$ in an LNF task $(V, P, A, I, G)$.

Compare Figure 6 with Figure 4. We deal with the expressions in constraints and effect right hand sides simply by inserting the respective $max$ values of the variables, and computing the respective outcome (recall that $exp(v)$ for an expression $exp$ and a variable value vector $v$ denotes the value of $exp$ when inserting the values $v$). The += effects are taken into account to obtain the $max_{t+1}$ values exactly as before, i.e., by adding their combined contributions to $max_t$ (except that the value of the right hand sides must now be computed using the $max_t$ values). The := effects are taken into account by determining, after all += effects have contributed to $max_{t+1}$, whether there is a := effect in the graph whose value, when inserting the $max_t$ values, is higher than the hitherto $max_{t+1}$ value. In this case, $max_{t+1}$ (for the respective variable) is updated to the maximum assignment possible.

The only part of the algorithm that becomes somewhat complicated, in comparison to the algorithm for restricted tasks, is the termination criterion. The difficult part is the computation of the mneed values, i.e., the values above which the variables can no longer contribute anything to a relaxed solution. These values can now depend on the state $s$ we start from. To derive the values, we start with the static (non state-dependent) notion of *solution-relevant* variables. A variable $v^i$ is solution-relevant if it either occurs in a numeric constraint, or in the right hand side $exp$ of an effect $(v^j, +=[:=], exp)$ on a solution-relevant variable $v^j$. Note that solution-relevance thus transfers transitively over the variables. We denote the set of solution-relevant variables with $rV$. For the state-dependent aspects of the relaxed task, we provide notation for the value that a variable $v^i$ must at least take on in a state $s$ in order to raise (or "support") the value of a positively weighted sum $exp = \sum_{j \in X} c^j * v^j + c$ above a constant $c'$.

$$\text{supv}^i(s, exp, c') := (c' - c - \sum_{i \neq j \in X} c^j * v^j(s)) \; / \; c^i$$

Of course, the support value $\text{supv}^i(s, exp, c')$ is only defined if $v^i \in v(exp)$, i.e., if $v^i$ is a part of the weighted sum. As the reader can easily convince him/herself, if we raise the value of $v^i$ in $s$ above $\text{supv}^i(s, exp, c')$ then we know that the value of $exp$ is at least $c'$. We use this concept to determine the point above which a variable $v^i$ contributes sufficiently to all constraints and effect right hand sides that it can contribute to. For constraints $(exp, \geq [>], 0)$ with $v^i \in v(exp)$ this point is reached with $v^i \geq \text{supv}^i(s, exp, 0)$ (then the constraint is fulfilled). For $+=$ effect right hand sides in $(v^j, +=, exp)$ with $v^i \in v(exp)$ and $v^j \in rV$ ($v^j$ may be needed) this point is reached with $v^i \geq \text{supv}^i(s, exp, 0)$ (the effect can then eventually increase $v^j$ to arbitrarily high values). As for $:=$ effect right hand sides, in an effect $(v^j, :=, exp)$ with $v^i \in v(exp)$ and $v^j \in rV$ the value of $v^i$ is sufficient if $v^i \geq \text{supv}^i(s, exp, \text{mneed}^j(s))$: then the effect is high enough to assign $v^j$ a sufficient value. The main complication here is that we want to use the supv values to define the mneed values so our definition for $:=$ effects is recursive. That does not constitute a problem given our assumption that the $:=$ effects are acyclic. In effect, the recursion is guaranteed to terminate. Altogether, the definition is the following.

$$\text{mneed}^i(s) := \max \begin{cases} -\infty, \\ \{\text{supv}^i(s, exp, 0) \mid (exp, \geq [>], 0) \in v(G) \cup \bigcup_{a \in A} v(pre(a)), v^i \in v(exp)\}, \\ \{\text{supv}^i(s, exp, 0) \mid (v^j, +=, exp) \in \bigcup_{a \in A} v(eff(a)), v^i \in v(exp), v^j \in rV\}, \\ \{\text{supv}^i(s, exp, \text{mneed}^j(s)) \mid (v^j, :=, exp) \in \bigcup_{a \in A} v(eff(a)), v^i \in v(exp), v^j \in rV\} \end{cases}$$

Note that, with this definition, the variables with $\text{mneed}^i(s) = -\infty$ are the variables that are not solution-relevant.

**Theorem 3** *Assume a linear numeric task $(V, P, A, I, G)$ that is in LNF and has acyclic $:=$ effects. Assume a state $s$. If the algorithm depicted in Figure 6 fails, then there is no relaxed plan for $(V, P, A, s, G)$.*

The main proof idea is, as before, this: if the algorithm fails at a layer $m$ then the termination condition will hold true at all later layers. The argument concerning the mneed$(s)$ values follows what is outlined above. The full details are a bit lengthy. See Appendix A.

As discussed before for the restricted language, the number of graph layers built before termination is finite – eventually, all variables either do not increase or reach their finite mneed values – but can be exponential in the encoding length of the task. Again, one could implement a provably polynomial algorithm along the lines of the method used in the proof to Theorem 2, and again it is debatable whether such an implementation would, for realistic examples, achieve any significant performance improvements over the existing implementation.

It is interesting to consider the role that the inverted variables – as introduced by Metric-FF during LNF pre-processing, see Section 4.3 – play in the relaxed planning graph process described above. Estimating the maximum value of an inverted variable is the same as estimating the minimum value of the respective original variable. More precisely, in Figure 6, if $v^j$ is the inverted variable to $v^i$ then $(-1) * max_t^j$ is, for all $t$, an optimistic approximation of the minimum value that $v^i$ can take on after $t$ steps: the value that results when one ignores all *increasing* effects on $v^i$, and is optimistic about the *decreasing* effects. In this sense, the introduction of the inverted variable $-v^i = v^j$ can be viewed as a way of informing the relaxed planner of where, in the numeric constraints and effects, to use the minimum or the maximum possible value of $v^i$, when computing an optimistic approximation of these maximum and minimum values.[14]

We now focus on relaxed plan extraction. As justified by Theorem 3, this is invoked only if the relaxed planning graph succeeds in reaching the goals. Also as before, the information that the graph provides are the levels of all actions, propositions, and numeric goals. For actions and propositions the definitions stay the same, for numeric goals $(exp, \geq [>], 0)$ the level is the graph layer $t$ where the goal can first be achieved, i.e., where $exp(max_t) \geq [>]0$ holds the first time. An outline of the plan extraction mechanism is shown in Figure 7.

Compared to the algorithm for restricted tasks, shown in Figure 5, the novelties in Figure 7 are that complex numeric goals get split up into goals for the individual variables, that effect right hand sides are forced to have a sufficiently high value, and that := effects are handled. The first issue, given a numeric goal $(exp, \geq [>], 0)$, is dealt with simply by constraining all variables $v^i \in v(exp)$ to take on their respective *max* value. Similarly, effect right hand sides in $(v^i, := [+=], exp)$ are forced to be sufficiently high by requiring all $v^j \in v(exp)$ to take on the respective *max* value. The := effects are taken into account as an alternative way of achieving a numeric goal $(v^i, \geq [>], c) \in v(G_t)$. If there is an effect $(v^i, :=, exp)$ with sufficiently high value, $exp(max_{t-1}) \geq [>]c$, then the respective action is selected. Otherwise a set of actions with += effects is selected in a similar fashion as for restricted tasks. As in Figure 5, when uniting sets of numeric goals that both contain a constraint on the same variable $v^i$, the stronger one of both constraints is taken. It is easy to see that, upon termination, the selected actions can be used to form a relaxed plan for the state at hand.

**Theorem 4** *Assume a linear numeric task $(V, P, A, I, G)$ that is in LNF and has acyclic := effects. Assume a state $s$ for which the algorithm depicted in Figure 6 reaches the goals. The actions selected by the algorithm depicted in Figure 7 form a relaxed plan for $(V, P, A, s, G)$.*

The (straightforward) proof can be found in Appendix A. We conclude this section with two additional remarks. One thing that might also have occurred to the reader is that one does not necessarily need to support a goal $(exp, \geq [>], 0)$ by requiring all $v^i \in v(exp)$ to take on the maximum possible value. Weaker requirements might already be sufficient. The same holds true for effect right hand sides. One might be able to find shorter relaxed plans by using some simple heuristics at these points. It also seems plausible that the algorithms specified here will work for *any* strictly monotonic task that uses only += effects and acyclic := effects, assuming the mneed value computation is modified appropriately. Exploring this idea for richer language classes is left open as a topic for future work. It is also left open if and how *= effects and /= effects could be taken into account.

---

14. This insight has been pointed out to the author by David Smith in a comment on the submitted version of this article. Optimistically estimating maximum and minimum variable values, or more generally multiple variable values, is an alternative viewpoint to the monotonicity paradigm we explore here. Investigating the alternative viewpoint in more depth is an open topic.

**for** $t := 1, \ldots, finallayer$ **do**
$\quad p(G_t) := \{g \in p(G) \mid level(g) = t\}$
$\quad v(G_t) := \{(v^i, \geq [>], max_t^i) \mid (exp, \geq [>]0) \in v(G), level(exp, \geq [>], 0) = t, v^i \in v(exp)\}$
**endfor**
**for** $t := finallayer, \ldots, 1$ **do**
$\quad$ **for all** $g \in p(G_t)$ **do**
$\quad\quad$ select $a$, $level(a) = t - 1$, $g \in p(eff(a))^+$
$\quad\quad$ **for all** $p \in p(pre(a)), (exp, \geq [>], 0) \in v(pre(a))$ **do**
$\quad\quad\quad p(G_{level(p)}) \cup = \{p\}$
$\quad\quad\quad v(G_{level(exp, \geq [>], 0)}) \cup = \{(v^i, \geq [>], max_{level(exp, \geq [>], 0)}^i) \mid v^i \in v(exp)\}$
$\quad\quad$ **endfor**
$\quad$ **endfor**
$\quad$ **for all** $(v^i, \geq [>], c) \in v(G_t)$ **do**
$\quad\quad$ **if** $\exists a, level(a) = t - 1, (v^i, :=, exp) \in v(eff(a)), exp(max_{t-1}) \geq [>]c$ **then**
$\quad\quad\quad v(G_{t-1}) \cup = \{(v^j, \geq, max_{t-1}^j) \mid v^j \in v(exp)\}$
$\quad\quad\quad$ /* introduce $a$'s preconditions as above */
$\quad\quad$ **else**
$\quad\quad\quad$ **while** $max_{t-1}^i \not\geq [\not>]c$ **do**
$\quad\quad\quad\quad$ select $a, level(a) = t - 1, (v^i, +=, exp) \in v(eff(a)), exp(max_{t-1}) > 0$
$\quad\quad\quad\quad\quad a$ not previously selected in this while-loop
$\quad\quad\quad\quad c := c - exp(max_{t-1})$
$\quad\quad\quad\quad$ /* introduce max constraints for all vars in $exp$ as above */
$\quad\quad\quad\quad$ /* introduce $a$'s preconditions as above */
$\quad\quad\quad$ **endwhile**
$\quad\quad\quad v(G_{t-1}) \cup = \{(v^i, \geq [>], c)\}$
$\quad\quad$ **endif**
$\quad$ **endfor**
**endfor**

Figure 7: Extracting a relaxed plan for a state $s$ in an LNF task $(V, P, A, I, G)$ (levels and $finallayer$ computed by the algorithm in Figure 1).

# 6. Metric-FF

This section details how the theoretical and algorithmic work described so far is used to implement the heuristic planning system Metric-FF. Section 6.1 specifies how the relaxed plan information is used to define the basic architecture of a planner that handles STRIPS plus linear tasks with acyclic := effects. We then describe extensions that are integrated into the system: Section 6.2 explains how the extension to ADL is handled, Section 6.3 explains how flexible optimization criteria can be taken into account.

## 6.1 Basic Architecture

As in the STRIPS case, once we have the techniques for extracting relaxed plans, a state space planner is easily implemented. The given linear task is transformed into an LNF task using the algorithms described in Section 4.3. We define a heuristic function, a search strategy, and a pruning technique analogous to that used in the STRIPS version of FF, c.f. Section 2.3. All methods are straightforward adaptions of the STRIPS techniques. The heuristic function still estimates goal distance as the number of actions in the relaxed plan.

**Definition 11** *Assume a linear numeric task* $(V, P, A, I, G)$ *that is in LNF and has acyclic* := *effects, and a state* $s$. *The Metric-FF heuristic value* $h(s)$ *for* $s$ *is defined as follows. If the algorithm depicted in Figure 6 fails,* $h(s) := \infty$. *Otherwise,* $h(s) := \sum_{t=1}^{finallayer} |A_t|$ *where* $A_t$ *is the set of actions selected at layer* $t$ *by the algorithm depicted in Figure 7.*

The search strategy remains exactly the same, namely enforced hill-climbing as depicted in Figure 3. The only difference lies in the the way we avoid repeated states. In the STRIPS case, this is a simple hash table lookup procedure. The straightforward adaption would be to store all visited states $s$, and cut out a new state $s'$ if an identical state $s$ has been visited before. We can, however, derive a weaker cutoff criterion that has an important performance impact in certain situations. It might be that $s'$ differs from $s$ only in that some solution-irrelevant numeric variables have other values. For example, the only difference between $s$ and $s'$ might be that in $s'$ more execution time has been spent. If we expand $s'$ then iteratively we might end up with an infinite sequence of succeeding states that do nothing but increase execution time (this phenomenon can be observed in various benchmark domains). We can avoid such phenomena by cutting out new states $s'$ that are *dominated* by a stored state $s$. Given a task $(V, P, A, I, G)$, a state $s'$ is dominated by a state $s$ if the propositions in $s$ and $s'$ are the same, and for all $v^i \in V$, either $v^i$ is not solution-relevant, $v^i \in V \smallsetminus rV$, or $v^i(s') \leq v^i(s)$ holds.[15] If $s'$ is dominated by $s$, and the task at hand is monotonic in the sense of Definition 8, then all action sequences that achieve the goal starting from $s'$ do the same starting from $s$.

**Proposition 5** *Assume a numeric task* $(V, P, A, I, G)$ *that is monotonic. Assume two states* $s$ *and* $s'$. *If* $s'$ *is dominated by* $s$ *then, for all action sequences* $P \in A^*$, *if* $result(s', P) \models G$ *then* $result(s, P) \models G$.

**Proof:** Say $P = \langle a_1, \ldots, a_n \rangle$ is an action sequence such that $result(s', P) \models G$ holds. We show that, for all solution-relevant variables $v^i \in rV$ and for all $0 \leq j \leq n$, $v^i(result(s', \langle a_1, \ldots, a_j \rangle)) \leq v^i(result(s, \langle a_1, \ldots, a_j \rangle))$ holds. This proves the proposition: the variables in the goal constraints are all in $rV$, the goal constraints are monotonic (Definition 8 condition (1)), and the goal constraints are fulfilled in $result(s', P)$. So with the claim above they are also fulfilled in $result(s, P)$. We prove the claim on the solution-relevant variable values by induction over $j$. Base case $j = 0$: by prerequisite, $v^i(s') \leq v^i(s)$ holds for all $v^i \in rV$. Inductive case $j \to j+1$. First, the preconditions of $a_{j+1}$ are fulfilled in $result(s, \langle a_1, \ldots, a_j \rangle)$ due to the same argument as used for the goal constraints above. Second, all variables that are contained in effect right hand sides on solution-relevant variables are themselves solution-relevant by definition so the induction hypothesis holds for them. This proves the claim with monotonicity of numeric effects (Definition 8 condition (2)). $\square$

---

15. Recall the definition of the solution-relevant variables $rV$, given in Section 5.2: all variables that occur in a numeric constraint, or in the right hand side $exp$ of an effect $(v^j, ass, exp)$ on a solution-relevant variable $v^j$.

LNF tasks are monotonic. So with Proposition 5, if there is a solution plan from $s'$, then there is a solution plan from $s$. Thus cutting $s'$ out of a search space that already contains $s$ is solution preserving. Consequently, during each search iteration performed by enforced hill-climbing, our implementation keeps a hash table of states visited in that iteration, and skips a new state if it is dominated by at least one of the visited states.[16] As indicated above, in various benchmark examples this prevents the planner from looping when the new states do nothing but increase the value of some solution-irrelevant variable like execution time.[17]

To extend our STRIPS pruning technique, helpful actions now are all those actions that can support either a propositional or a numeric goal at the lowest layer of the relaxed planning graph.

**Definition 12** *Assume a linear numeric task $(V, P, A, I, G)$ that is in LNF and has acyclic := effects, and a state $s$ for which the algorithm depicted in Figure 6 reaches the goals. The set of helpful actions $H(s)$ for $s$ is defined as*

$$H(s) := \{\, a \in A \mid p(\text{eff}(a))^+ \cap p(G_1) \neq \emptyset \vee$$
$$\exists (v^i, \geq [>]c) \in v(G_1) : \exists (v^i, :=, \exp) \in v(\text{eff}(a)) : \exp(v(s)) \geq [>]c \vee$$
$$\exists (v^i, \geq [>]c) \in v(G_1) : \exists (v^i, +=, \exp) \in v(\text{eff}(a)) : \exp(v(s)) > 0 \,\},$$

*where $G_1$ is the set of sub-goals constructed at layer 1 by the algorithm depicted in Figure 7.*

Supporting a numeric goal here means: for := effects, that the right hand side of the effect is sufficient to fulfill the goal; for += effects, that the respective right hand side expression is greater than 0. Note that the right hand side value of an effect at the lowest layer of the relaxed planning graph is exactly its value in the state $s$ at hand. During a search iteration in enforced hill-climbing, when expanding a state $s$, only the states generated by the actions in $H(s)$ are included into the search space. Note that states $s$ where the relaxed planning graph does not reach the goals have $h(s) = \infty$ so do not get expanded anyway.

As in STRIPS, the algorithm can fail if either enforced hill-climbing gets trapped in a dead end state or helpful actions pruning cuts out important states. We have observed that helpful actions pruning is too severe in some numeric domains. So, in case enforced hill-climbing fails we try again with the pruning technique turned off, i.e., we continue the hill-climbing procedure from the point of failure without pruning. If this fails too, then like in STRIPS we employ a safety net solution: a complete greedy best-first strategy trying to solve the task from scratch. This strategy expands all search nodes by increasing order of goal distance estimation. New states are cut out if they are dominated by an already visited state.

---

16. More precisely, a new state $s'$ is skipped only if there is a dominant visited state $s$ *in the same hash entry*. If the value of all solution-relevant variables is the same in $s'$ and $s$ (like when only execution time has increased), then our implementation ensures that this is the case. Otherwise it is a matter of chance. It is an open question how the visited states could be indexed in order to provide a fast exact answer to the query whether they contain a dominant state or not.

17. What we have here is a consequence of the undecidability of numeric planning (Helmert, 2002), which can be observed even in seemingly benign benchmarks. In a finite state space we of course would not run the risk of entering an infinite loop.

## 6.2 ADL

ADL (Pednault, 1989) goes beyond STRIPS in that it allows, in action preconditions and the goal, arbitrary equation-free first-order logical formulae, and actions with conditional effects – effects that only occur when their effect condition holds true. The effect condition can be an arbitrary (equation-free) first-order logical formula. In the numeric setting, the effects can contain updates on numeric variables. The numeric constraints can now appear at any point in a logical formula where a logical atom is allowed.

Like the previous FF version (Hoffmann & Nebel, 2001), Metric-FF compiles quantifiers and disjunctions away in a pre-processing phase. Metric-FF does not compile conditional effects away. So Metric-FF's internal language differs from STRIPS (with numeric constraints and effects) only in that actions can have conditional effects, where the effect conditions are conjunctions of propositions (and numeric constraints). The reason why ADL is compiled into this language is that the heuristic algorithms (i.e., the relaxed planning graph) can be implemented very efficiently for this more restricted language format. The compilation can be exponentially costly in general but is feasible when, as one might expect in the formulation of a realistic planning scenario, the logical formulae are not overly complex. The reason why conditional effects are not also compiled away (which could be done in principle) is that, as Nebel (2000) proved, this would imply another exponential blow up given we want to preserve solution length. Fortunately the conditional effects can easily be dealt with so there is no need to compile them away. In the following, we give a brief overview of the compilation process, and of the extended heuristic function implementation. Except for the heuristic function, the only thing that must be adapted is the state transition function, which is conceptually trivial.

The compilation process is largely an implementation of ideas that have been proposed by Gazen and Knoblock (1997), as well as Koehler and Hoffmann (2000b). The extensions to handle numeric constructs are all straightforward. The process starts with the usual planner inputs, i.e., with a set of parameterized operator schemata, an initial state, and a goal formula. The compilation works as follows.

1. Determine predicates and numeric functions that are static in the sense that no operator has an effect on them. Such predicates and functions are a common phenomenon in benchmark tasks. Examples, in a transportation context, would be the connections between locations as given by a static (connected ?l1 ?l2) predicate, or the distances between locations as given by a static (distance ?l1 ?l2) function. Static predicates and functions are recognized by a simple sweep over all operator schemata.

2. Transform all formulae into quantifier-free DNF. This is subdivided into three steps:

   (a) Pre-normalize all logical formulae. Following Gazen and Knoblock (1997), this process expands all quantifiers, and translates negations. We end up with formulae that consist of conjunctions, disjunctions, and atoms containing variables (where the atoms can be numeric constraints).

   (b) Instantiate all parameters. This is simply done by instantiating all operator and effect parameters with all type consistent constants one after the other. The process makes use of knowledge about static predicates, in the sense that the instantiated formulae can often be simplified (Koehler & Hoffmann, 2000b). For example, if an instantiated static predicate $(p\ \vec{a})$ occurs in a formula, and that instantiation is not contained in the initial state, then $(p\ \vec{a})$ can be replaced with "false". As another example, if both sides of a numeric constraint are static then the constraint can be replaced with either "true" or "false".

(c) Transform formulae into DNF. This is postponed until after instantiation, because it can be costly, so it should be applied to the smallest formulae possible. In a fully instantiated formula, it is likely that many static predicate occurrences (constant constraint occurrences) can be replaced by "true" or "false", resulting in a much simpler formula structure.

3. Finally, if the DNF of any formula contains more than one disjunct, then the corresponding effect, operator, or goal condition gets split up in the manner proposed by Gazen and Knoblock (1997).

When all the logical constructs have been normalized, the numeric constructs in the task are transformed into LNF in a manner analogous to the process described in Section 4.3. Integrating conditional effects into the relaxed planning process is an easy matter. The relaxed planning graph differs from its STRIPS counterpart only in that it now keeps track of the graph layers at which an action's effects first become applicable. The relaxed plan extraction process differs from its STRIPS counterpart only in that it now selects supporting *effects* for the propositional and numeric goals.

## 6.3 Optimization Criteria

In PDDL2.1, the user can specify an optimization criterion for a task. The criterion consists of an arbitrary numeric expression together with a keyword "maximize" or "minimize" saying whether higher or lower values of the expression are preferred. The semantics are that a solution plan is optimal iff the state it leads to is a maximal / minimal goal state with respect to the optimization expression. Metric-FF supports, run in "optimization mode", a somewhat more restrictive form of optimization. It accepts the optimization criterion only if the criterion can be transformed, according to a certain schema, into additive action cost minimization. The heuristic cost of a state is then the summed up cost of the actions in the respective relaxed plan, and search is a standard weighted $A^*$ where the weights can be set via the command line. Note that this methodology can not give a guarantee on the quality of the returned solution as the heuristic function is not provably admissible. The methodology is an obvious option given that the cost of a relaxed plan (in an additive setting) gives us a remaining cost estimation technique for free. It is an open question how more general optimization criteria can be dealt with. In the following, we describe our implemented methodology in a little more detail. We start with the STRIPS setting, then outline the changes made in the extension to ADL.

Metric-FF rejects the optimization expression if it is not linear. Otherwise, if the optimization keyword is "maximize" then the expression is multiplied by $-1$ so minimization is required. The expression is then brought into LNF, $\sum_{j \in X} c^j * v^j$ (the constant part can obviously be skipped). With this notation, the optimization criterion is accepted (only) if all action effects on variables $v^j \in X$ increase the optimization expression value by a constant, i.e., if all effects on $v^j \in X$ are of the form $(v^j, +=, c)$ where $c \in \mathbf{Q}$, $c \geq 0$. For each action $a$, the cost of the action is then defined as

$$cost(a) := \sum_{(v^j, +=, c) \in v(e\!f\!\!f(a)), v^j \in X} c^j * c,$$

i.e., as the sum of all increases in cost variables for the action, multiplied by their weight in the optimization expression. The cost of an action sequence or set is the sum of the individual costs. It is easy to see that, in this setting, finding a goal state that minimizes the optimization expression value is equivalent to finding a plan with minimal cost. The search algorithm we then use is, as stated above, a standard weighted $A^*$ algorithm implementing a best-first search on the function $f(s) = w_g * g(s) + w_h * h(s)$ where $g(s)$ is the cost of the search path that leads to $s$, $h(s)$ is the remaining cost estimate (i.e., the cost of the relaxed plan from $s$), and the weights $w_g$ and $w_h$ can be given in the command line. Since the remaining cost estimate is in general not admissible, the first plan found is not guaranteed to be optimal. But one would expect that empirically better plans can be found. We will see below that this is, in fact, the case in some of the IPC-3 testing domains.

In ADL, the cost of an action in a state is the sum of the costs of all effects that appear, the cost of an action sequence is the sum of the costs of the actions in the respective states, and minimizing the optimization expression is as before equivalent to minimizing plan cost. Estimating the remaining cost by means of a relaxed plan becomes somewhat less obvious, since a choice has to be made on which effect costs are counted for the result. There are effects that have been selected to support logical or numeric goals during relaxed plan extraction, and there are effects that will get triggered when actually executing the relaxed plan. We have chosen to only count the costs of the former effects. The heuristic search algorithm remains exactly the same as in the STRIPS case.

## 7. Competition Results

We briefly examine the IPC-3 competition data relevant to Metric-FF. The competition featured domains spanning the whole range from STRIPS to PDDL2.1 level 3, which permits a combination of logical, numeric, and temporal constructs. FF participated in the STRIPS domains and in the numeric domains, demonstrating very competitive performance. We only discuss the data for the numeric domains. A discussion of the STRIPS results can be found in the competition overview article by Long and Fox (2003).

There were six numeric domains used in the competition. For each of these domains, we include a figure showing runtime curves, and discuss relative (runtime and solution quality) performance in the text. Like FF, the MIPS and LPG systems could be configured to either favor speed or quality, i.e., to either find some plan as fast as possible or to search for a good plan in the sense of the optimization criterion. To make the graphs readable, we only show the runtime curves of those planners that favor speed. We discuss the solution quality behavior of these planners in terms of plan length, i.e., number of steps. Note that these planners do not take account of the optimization criterion anyway. For the planners that favor quality, we discuss their runtime and solution quality behavior in the text. Given that the optimization mode in Metric-FF is only a preliminary implementation, we keep the discussions short. We also give only brief descriptions of the domain semantics. More details on these can be found in the overview article (Long & Fox, 2003). We focus on the six domains in turn, then give a short summary of Metric-FF's performance.

## 7.1 Depots

The *Depots* domain is a combination of the well-known *Logistics* and *Blocksworld* domains. Objects must be transported with trucks as in *Logistics*, and must then be arranged in stacks as in *Blocksworld*. The numeric constructs define fuel consumption for trucks and the hoists that lift the objects (in order to stack them somewhere). Objects have weights and the sum of the weights of the objects loaded onto a truck at any time must be lower than or equal to that truck's capacity. Figure 8 shows the runtime data on the 22 *Depots* instances used in the competition.

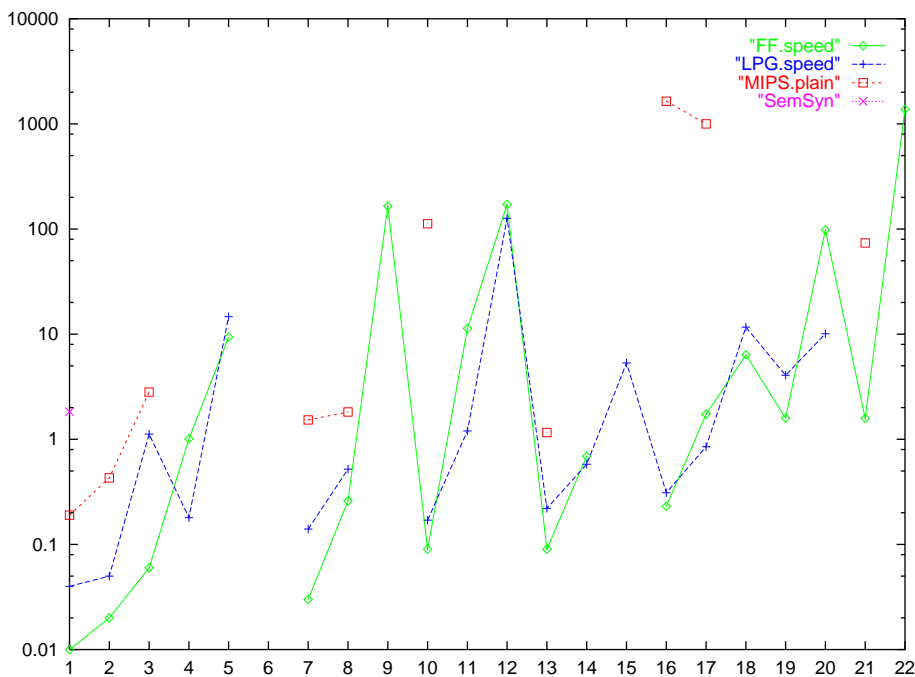

Figure 8: Runtime curves on *Depots* instances for the planners favoring speed. Time is shown on a logarithmic scale, instance size scales from left to right.

The four planners participating in the numeric version of *Depots* were Metric-FF, LPG, MIPS, and SemSyn. At the time of writing, no paper on the numeric version of any of these planners is published. for LPG and MIPS, the reader is referred to the respective articles to appear in this same JAIR special issue (Gerevini et al., 2003a; Edelkamp, 2003). As Figure 8 shows, SemSyn can only solve the single smallest instance, and MIPS solves 10 instances scattered across the whole set. Metric-FF and LPG solve most of the instances and exhibit similar behavior. Metric-FF is the only planner that can solve the two largest instances. As stated above, we only show the curves for those configurations favoring speed. In the competition data, this version of Metric-FF is called "FF.speed", this version of LPG is called "LPG.speed", and this version of MIPS is called "MIPS.plain".

To assess relative plan quality behavior (i.e., plan length or minimization expression value), we computed quotients as follows. Given planners A and B, measure, for all instances solved by both planners, A's plan quality divided by B's plan quality. Compute the average quotient. At points where we need an absolute measure of comparison between the participating planners in a domain, we set the planner B to a hypothetical "Best-of" planner whose data is obtained by selecting the best (i.e., lowest) results of all planners. The individual planners in the domain are then all ranked by comparing them to Best-of.

The data obtained concerning plan length in *Depots*, for the planners shown in Figure 8, is this. FF.speed's plans are on average 1.23 times as long as Best-of's plans, LPG.speed's plans are on average 1.25 times as long as Best-of's plans, and MIPS.plain's plans are on average 1.29 times as long as Best-of's plans. Thus plan lengths are roughly similar here. For the single instance that SemSyn solves, its plan has 5 steps while FF.speed's has 10 steps, and those of LPG.speed and MIPS.plain have 11 steps.

We next comment on the algorithms used in the planner versions favoring quality. In MIPS, similar to Metric-FF, in optimization mode the heuristic function becomes a kind of relaxed plan cost in an $A^*$ algorithm. In contrast, the LPG optimization method starts from the first plan, and then continues search for plans that are better. Metric-FF performs best-first search on the function $f(s) = w_g * g(s) + w_h * h(s)$. In the competition, the weights were set to $w_g = 1$ and $w_h = 5$. The quality version of MIPS is simply called "MIPS" in the competition data. To improve readability we call it "MIPS.quality" here, similar to the quality-favoring versions of Metric-FF and LPG, called "FF.quality" and "LPG.quality".

The optimization criterion in *Depots* is to minimize overall fuel consumption. For runtime, the quality versions of MIPS and LPG behave only slightly worse than the speed versions. In contrast, Metric-FF's quality version solves only the smallest 3 instances. For solution quality, the fuel consumption of FF.speed on the first 3 instances is 22, 33, and 35, while that of FF.quality is 22, 33, and 36. Thus no optimization effect is observable. On the same instances, MIPS.quality finds more costly plans (32, 63, and 44), and LPG.quality's plans are slightly better (22, 33, and 29). Across all instances, LPG.quality's plans consume, on average, 1.01 times the fuel that Best-of's plans consume, while that average value is 1.46 for MIPS.quality.

## 7.2 Driverlog

The *Driverlog* domain is a variation of *Logistics* where the trucks need drivers, and the underlying map is an arbitrary undirected graph (as opposed to the fully connected graphs in the standard version of the domain). Drivers can move on different paths than trucks. The numeric constructs specify the total time driven and walked. Figure 9 shows the runtime data for the 20 *Driverlog* instances used in the competition.

As in *Depots*, the participating planners were Metric-FF, LPG, MIPS, and SemSyn. Again, SemSyn solved only the smallest instance. LPG.speed is the only planner that solves all instances. FF.speed solves one more task than MIPS.plain (the respective data point is almost hidden behind "SemSyn" in the top right corner), and is roughly as fast as LPG.speed on the tasks that it solves. As for plan length, again none of the planners is clearly superior. The average quotients versus Best-of are: 1.34 for FF.speed, 1.44 for LPG.speed, and 1.21 for MIPS.plain. FF.speed's and LPG.speed's plan lengths are thus on average somewhat longer than those of MIPS.plain. The difference has no tendency to grow with instance size, though. On the single instance solved by SemSyn, SemSyn's plan has 3 steps while those of the other planners have 8.

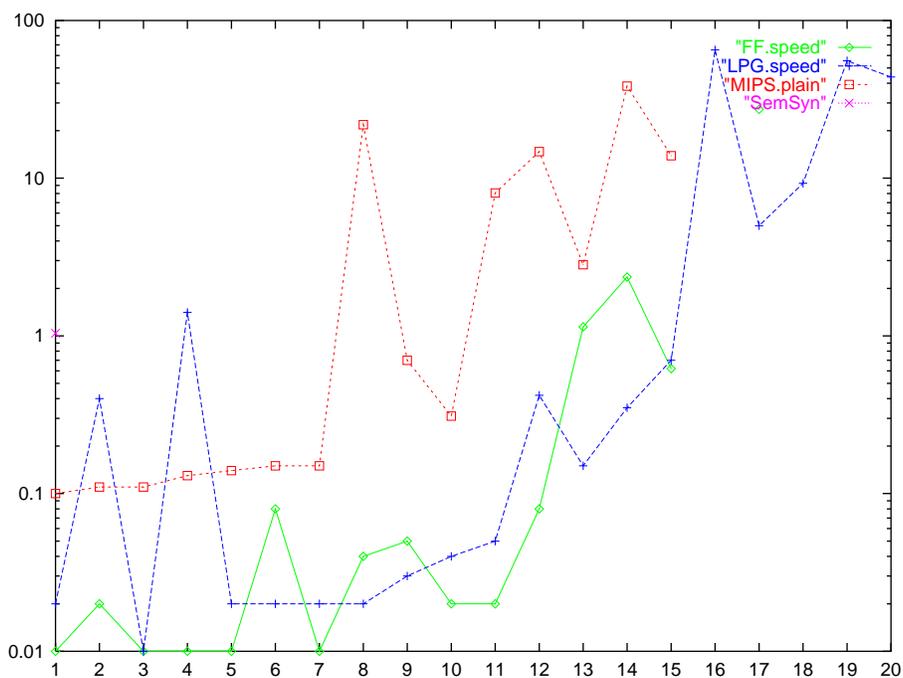

Figure 9: Runtime curves on *Driverlog* instances for the planners favoring speed. Time is shown on a logarithmic scale, instance size scales from left to right.

The optimization criterion in *Driverlog* is to minimize some (instance-specific) linear combination of total time, driven distance, and walked distance. FF.quality's runtime behavior is, as in *Depots*, a lot worse than that of FF.speed, solving only 5 of the smaller instances. The quality of the plans is slightly better, though, 0.94 times FF.speed's values on average. MIPS.quality and LPG.quality solve the same instances as their speed-favoring counterparts. The average comparison of LPG.quality to Best-of is 1.00 (precisely 1.000411), that of MIPS.quality is 1.31 – on a single instance, MIPS.quality's plan consumes less fuel (730 units) than LPG.quality's plan (736 units).

The competition also featured a version of *Driverlog* ("Hard-Numeric") where driving a truck consumes fuel proportional to the square of its load, and the criterion is to minimize an instance-specific linear combination of total time and fuel consumption. Interestingly, with this optimization criterion FF.quality is only slightly less efficient than FF.speed, solving the same instances as the speed-favoring version. We will come back to this phenomenon in the outlook, when we discuss the effect of optimization expressions on runtime performance. The overall runtime performance of all other planners is similar to that in the domain version described above. For the optimization expression, FF.quality's values are on average 0.77 times those of FF.speed (so an optimization effect can be observed). The comparison to Best-of is 1.59 for FF.quality, 1.007 for LPG.quality, and 1.72 for MIPS.quality.

### 7.3 Zenotravel

The *Zenotravel* domain, as used in the competition, is a rather classical transportation domain, where objects must be transported via airplanes. The planes use fuel, and can fly either slow or fast. Fast movement consumes more fuel. In the numeric version of the domain, the fuel level of a plane and the overall fuel usage are numeric variables. In addition, a numeric variable counts the passengers on board a plane, and fast movement is only allowed if the number of passengers is below a certain threshold. A refuel operator can be used to set the fuel level of a plane back to its maximum capacity. Without durations, the only difference between the effects of slow and fast flying lie in the higher fuel consumption, thus "fast" flying is a useless action. Figure 10 shows the runtime data on the 20 *Zenotravel* instances used in the competition.

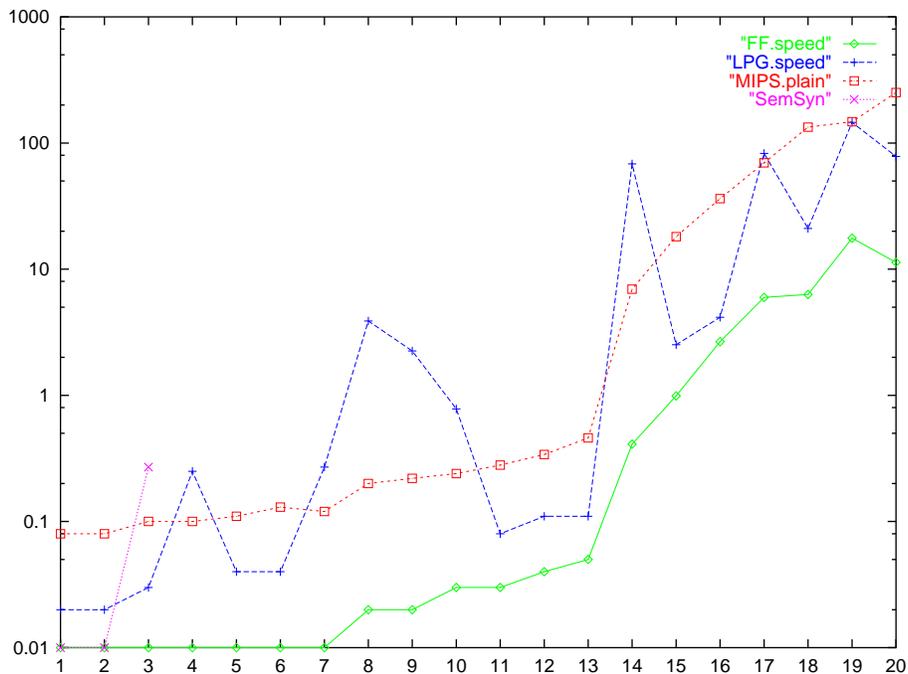

Figure 10: Runtime curves on *Zenotravel* instances for the planners favoring speed. Time is shown on a logarithmic scale, instance size scales from left to right.

Again, the participating planners were Metric-FF, LPG, MIPS, and SemSyn. Sem-Syn solves only the smallest three instances, the other planners solve the whole test set. FF.speed is an order of magnitude faster than LPG.speed and MIPS.plain. For plan length, FF.speed and MIPS.plain behave similarly, while LPG.speed finds somewhat longer plans. The quotient values versus Best-of are 1.28 for FF.speed, 1.45 for LPG.speed, and 1.22 for MIPS. When visualizing the data, one finds that the difference between LPG.speed's and FF.speed's plans grows with instance size. SemSyn, again, finds the best (shortest) plans for those instances that it solves. The quotients FF.speed versus SemSyn are 1.00, 2.72, and 4.50 on the three instances solved by SemSyn.

The optimization criterion in *Zenotravel* is to minimize some (instance-specific) linear combination of total time and fuel consumption. FF.quality's runtime behavior is worse than that of FF.speed, solving only the smaller half of the test set. MIPS.quality solves only the first 16 instances, LPG.quality solves all but the largest instance. The optimization criterion values of FF.quality are on average 0.82 times those of FF.speed, so an optimization effect can be observed. The quotient values versus Best-of are 1.51 for FF.quality, 1.39 for LPG.quality, and 1.14 for MIPS.quality.

### 7.4 Satellite

In *Satellite*, a number of Satellites must make a number of observations using their installed instruments. This involves turning the Satellites the right direction, switching the instruments on or off, calibrating the instruments, and taking images. In the numeric version of the domain, turning the Satellites consumes (non-replenishable) fuel, the images occupy data memory, and the Satellites have only limited data memory capacity. Figure 11 shows the runtime data on the 20 problem instances used in the competition.

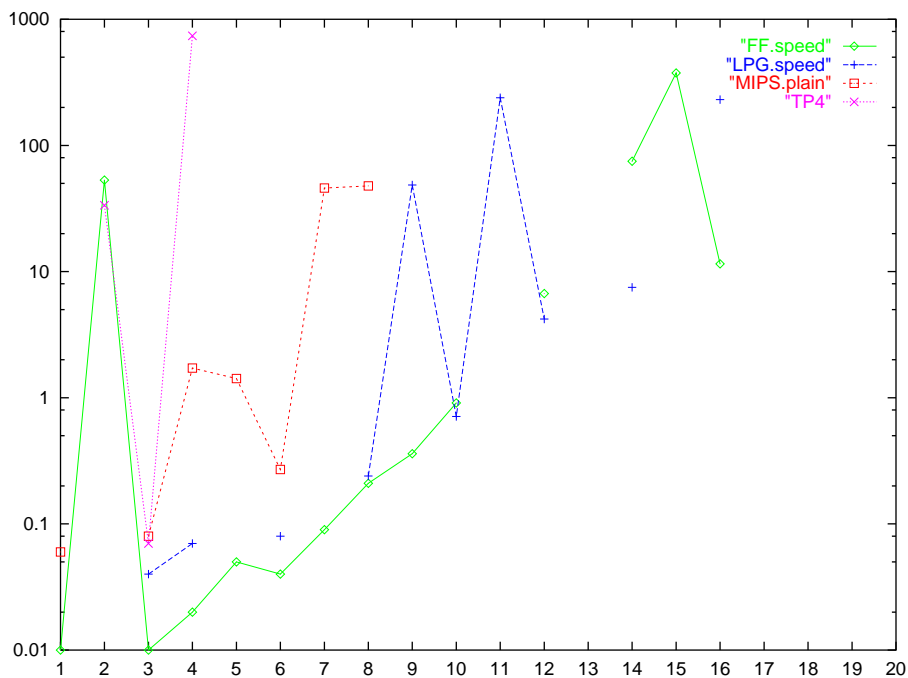

Figure 11: Runtime curves on *Satellite* instances for the planners favoring speed. Time is shown on a logarithmic scale, instance size scales from left to right.

In this domain, the participating planners were Metric-FF, LPG, MIPS, and TP4 (Haslum & Geffner, 2001). TP4 (which finds plans with optimal makespan) solves 3 of the smallest instances, MIPS.plain solves 7 of the smaller instances, LPG.speed solves 10 instances, and FF.speed solves 14. Note, though, that the instances that LPG.speed fails to solve but FF.speed solves are mainly the smaller ones. As for plan length, the quotients versus Best-of are 1.11 for FF.speed, 1.04 for LPG.speed and MIPS.plain, and 1.09 for TP4.

So plan lengths are roughly similar, but LPG.speed and MIPS.plain seem to have a slight advantage over FF.speed.

The optimization criterion in *Satellite* is to minimize overall fuel consumption. FF.quality's (MIPS.quality's) runtime behavior is a lot worse than that of FF.speed (MIPS.plain), solving only 2 (4) of the smallest instances. LPG.quality solves the same instances as LPG.speed. The fuel consumption of FF.speed on the 2 instances that FF.quality solves is 109 and 97. That of FF.quality is 109 and 83, so there is a slight optimization effect on one of the two instances. LPG.quality finds the best plans for all instances that it solves (thus the quotient versus Best-of is constantly 1.00), the comparison of MIPS.quality to Best-of is 2.54.

The competition also featured a version of *Satellite* ("Hard-Numeric") where there were no logical or numeric goals at all, and the optimization criterion was to maximize the amount of stored data (i.e., the memory occupied by the taken images). This is an example of an optimization criterion that can not be transformed into action costs in the sense explained in Section 6.3. The actions that take images have negative costs. Metric-FF thus rejects the optimization criterion and reports, for all instances, that they are trivially solved by the empty plan. Similarly, the plans returned by MIPS.plain are all empty. The MIPS.quality version finds non-trivial plans for the smaller half of the instances. For LPG there is no data in the competition results for this domain version.

## 7.5 Rovers

In *Rovers*, a number of planetary rovers must analyze a number of rock or soil samples, and take a number of images. This involves navigating the rovers, taking or dropping samples (rovers can only hold one sample at a time), calibrating the camera and taking images, and communicating the data to a lander. In the numeric version of the domain, all the activities decrease the energy available for the rover by a certain amount, and an energy recharge operator can be applied when the rover is located in a sunny spot. Figure 11 shows the runtime data on the 20 *Rovers* instances used in the competition.

The participating planners in this domain were Metric-FF, LPG, and MIPS. None of the planners can solve the whole test set, in fact LPG, which scales best, is the only planner that can solve most of the larger instances.[18] The smaller instances are solved quickly by all three participants. FF.speed might have a slight plan length advantage. The quotients versus Best-of are 1.02 for FF.speed, 1.26 for LPG.speed, and 1.19 for MIPS.plain.

The optimization criterion in *Rovers* is to minimize the number of recharge actions applied in the plan (i.e., the cost of recharging is 1, the cost of all other actions is 0). With this optimization criterion, FF.quality does not solve a single instance (we will return to this in the outlook). MIPS.quality and LPG.quality solve the same instances as their speed-favoring counterparts. LPG.quality's plan quality is 0 in all the 8 instances that MIPS.quality solves. MIPS.quality's plans contain 0 recharge actions in three cases, 1 recharge action in four cases, and 2 recharge actions in one case.

---

18. In the actual competition data, LPG failed to solve 8 of the instances due to an implementation bug. We show the corrected data provided by Alfonso Gerevini.

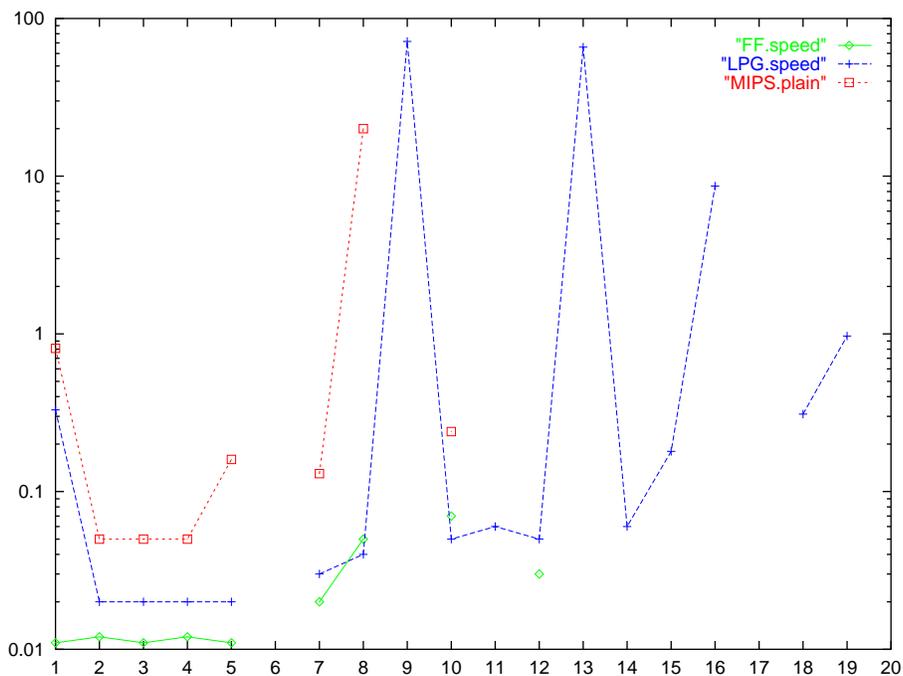

Figure 12: Runtime curves on *Rovers* instances for the planners favoring speed. Time is shown on a logarithmic scale, instance size scales from left to right.

## 7.6 Settlers

The *Settlers* domain is about building up an infrastructure in an unsettled area. The things to be built include housing, railway tracks, sawmills, etc. There are a lot of operator schemata encoding a complex building process. The raw materials, timber, stone, and ore, must first be felled, broken, or mined. One can then process timber into wood or coal, and process ore into iron. Carts, trains, or ships can be built to transport materials. One can combine materials to build docks, wharfs, rails, housing, etc. The encoding makes a more intensive use of numeric variables than the other domains. While in the other domains the numeric constructs mainly encode resource constraints and action costs, in *Settlers* the numeric variables play an active part in achieving the goal. Indeed, many of the operator schemata have no logical effects at all. For example, felling timber increases the amount of timber available at the respective location by one unit. Loading (unloading) a material unit onto (from) a vehicle is encoded by increasing (decreasing) the respective material availability in the vehicle while decreasing (increasing) the material's availability at the respective location. For building a housing unit at least one wood and stone unit must be available, resulting in increased housing units and decreased wood and stone units. With the numeric variables playing such an active role in the domain encoding, *Settlers* is a very interesting benchmark for numeric planners. Figure 11 shows the runtime data on the 20 *Settlers* instances used in the competition.

Only Metric-FF and MIPS (in the versions that favor speed) were able to solve some of the *Settlers* instances. LPG could not participate in this domain because some operators make use of universally quantified effects, which LPG does not support. MIPS.plain solves only a single instance while FF.speed solves the 6 smallest instances. It should be noted here that the instances in this example suite appear to be rather large. FF.speed's plans on the 6 smallest instances contain 53, 26, 102, 67, 74, and 81 actions respectively. For comparison, in all of the other domains except *Depots* FF.speed's longest plan in the first 6 instances contains 26 steps. In *Depots* the numbers are 10, 15, 35, 34, 75, the 6th instance isn't solved by any planner. The plan that MIPS.plain finds for the second smallest instance contains 36 steps (as stated above FF.speed's plan for this instance contains 26 steps). No planner favoring quality solved any of the *Settlers* instances.

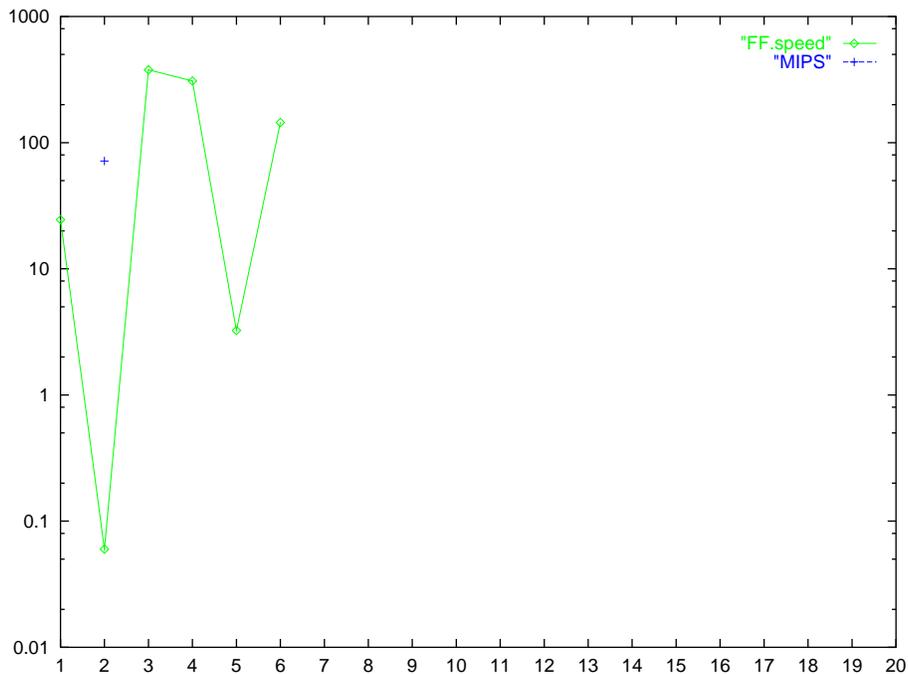

Figure 13: Runtime curves on *Settlers* instances for the planners favoring speed. Time is shown on a logarithmic scale, instance size scales from left to right.

## 7.7 Performance Summary

In their speed-favoring configurations, Metric-FF and LPG perform the best, both in terms of runtime and solution length. For runtime, in *Driverlog* and *Rovers*, LPG scales better (solves more instances). In *Zenotravel*, Metric-FF scales better (an order of magnitude advantage in runtime). In *Settlers*, LPG could not be run, but Metric-FF can solve some rather large instances. In *Depots* and *Satellite*, there is a slight advantage for Metric-FF, which solves a few more instances. MIPS lags behind both Metric-FF and LPG in all the domains except *Zenotravel* where it scales roughly similar to LPG. As for solution length, this is roughly similar for LPG and Metric-FF in all of the domains except *Zenotravel*, where LPG's plans are longer. In *Satellite* there might be a slight advantage for LPG, and in *Rovers* there might be a slight advantage for Metric-FF. The plan lengths of MIPS are roughly similar to those of Metric-FF across all the domains.

The results for Metric-FF in optimization mode, FF.quality, are less satisfying, at least as far as runtime behavior is concerned. FF.quality does not solve a single instance in *Rovers* and *Settlers*, and it solves only very few instances in *Depots*, *Driverlog* (with the normal, i.e. not "hard" optimization expression), and *Satellite*. FF.quality's runtime behavior is reasonably good only in *Zenotravel* and *Driverlog* (with the "hard" optimization expression). The solution quality behavior is mixed. In most cases it can be observed that FF.quality's plans are better in the sense of the optimization criterion than FF.speed's plans are. Better plan quality is clearly observable in *Driverlog* (with the "hard" optimization expression) and *Zenotravel*. It is also observable in *Driverlog* (with the normal optimization expression) and *Satellite*, although only a small number of instances were solved in these domains. Compared to LPG.quality and MIPS.quality, FF.quality is the only quality-favoring planner here that shows dramatically worse runtime behavior than its speed-favoring counterpart. The reasons for that must lie in the algorithmic differences between the systems, concerning the way they treat optimization expressions. In the outlook we speculate on the reasons for FF.quality's poor runtime behavior, and what might be done about it.

## 8. Conclusion and Outlook

We have presented a natural extension of a popular heuristic technique for STRIPS – ignoring delete lists – to numeric planning. The straightforward implementation of Metric-FF based on the technique was one of the two best performing numeric planning systems at IPC-3.

Let us summarize the contributions of this work in a little more detail. The most important contribution is the "monotonicity" idea, i.e., a numeric framework in which the main STRIPS concepts (pre/goal-conditions, add lists, and delete lists) translate very naturally to the numeric concepts (monotonic constraints, increasing effects, and decreasing effects). The monotonicity idea might be useful in many other contexts beside the specific heuristic planner implementation we focus on in this article (some ideas on that are given in the outlook below). In the heuristic context considered here, we have:

- Abstracted the desirable properties (admissibility, basic informedness, and polynomiality) that ignoring delete lists has as a relaxation in STRIPS.

- Defined a natural extension of this relaxation to the numeric case and provided sufficient criteria to identify numeric tasks where the relaxation preserves the desirable properties.

- Defined a subset of PDDL2.1 level 2, linear tasks, where the sufficient criteria can be achieved by a pre-processing technique.

- Defined algorithms that solve relaxed tasks in this language and thus provide a heuristic function.

- Implemented a straightforward extension of FF, and a first technique that takes user-specified optimization criteria into account. The FF extension (FF.speed) shows reasonable performance across a number of benchmark domains. Specifically it performed best, together with LPG, in the numeric track of the 3rd International Planning Competition, both in terms of runtime and solution length.[19] The runtime behavior of the optimization technique (FF.quality) is unsatisfying, but plan quality improvements can be observed.

Various research topics have been left open:

- The background theory given in Section 4.2 provides only sufficient criteria for the numeric relaxation to be adequate. The question is, are there weaker sufficient criteria, and can one come up with a complete analysis (i.e., find the exact borders beyond which ignoring decreasing effects is no longer adequate)? Also, how do the identified borderlines translate, syntactically, to the mathematical constructs allowed in PDDL2.1 level 2?

- The pre-processing algorithm given in Section 4.3 (transforming linear tasks into LNF tasks) is defined for linear tasks only. Can it be extended to richer language classes? Similarly, the algorithms given in Section 5.2 only work for LNF. Is there an easy extension to richer language classes?

- As mentioned in Section 4.3, various kinds of numeric effects can easily be translated into each other (e.g., := effects into += effects or vice versa), but the respective translations behave differently in the relaxation. Can one identify problem classes where one or the other formulation yields better heuristic performance?

- The current optimization technique, FF.quality, is restricted to optimization criteria that can be transformed into action cost minimization according to a certain simple translation schema. How can more general optimization criteria be handled?

- We have seen that the runtime performance of FF.quality is unsatisfying. There appears to be some interaction (as exemplified by the two different quality metrics in *Driverlog*) between the form of the optimization (i.e., the action cost minimization) expression and runtime behavior. An explanation for this might be the degree of "goal-directedness" of the minimization expression. Intuitively, a minimization expression is goal-directed if it is closely correlated with goal distance, i.e., the lower the expression value the nearer the goal and vice versa. The maximally goal-directed minimization expression is the goal distance itself (i.e., "total-time" in our sequential framework). In contrast, the minimization expression in *Rovers*, number of recharge operations, is only very loosely connected with goal distance. It would be worthwhile to come up with a good formal notion of goal-directedness, and to investigate its connection with runtime performance (in *Driverlog* the connection is less obvious than in *Rovers*). On the more practical side, algorithms remain to be found that show better performance no matter what the form of the optimization expression is. One option is to always integrate, to some extent, the current goal distance estimate (i.e., the length of the relaxed plan in our case) into the remaining cost estimation. Another option is to use different search schemes. A branch-and-bound like approach appears possible (first find some plan quickly then use the cost of this plan as an upper bound during further exploration of the search space).

---

19. Note that one can easily imagine domains where relaxed plans in the way Metric-FF uses them would likely yield no good heuristic information. As an example, consider the 15-puzzle, with numeric variables encoding the positions of the tiles. In this situation, there is a large degree of interaction between the numeric variables, and relaxed plans will presumably not be able to capture this interaction.

It would be exciting to explore the impact of the monotonicity idea, i.e., the correspondence that it brings between pre/goal-conditions and monotonic constraints, add lists and increasing effects, as well as delete lists and decreasing effects, in different contexts of planning research. Examples that spring to mind are other heuristic approaches, Graphplan-based numeric planning, or goal ordering techniques. To stimulate the imagination of the reader:

- It seems likely that similar methods can be used in other heuristic approaches that relax the task by ignoring the delete lists. For example, our techniques can presumably be adapted to heuristic estimators in the partial order framework used in RePOP (Nguyen & Kambhampati, 2001), yielding a heuristic numeric partial-order planner. Also, it appears feasible to integrate our techniques into Sapa's (Do & Kambhampati, 2001) heuristic function, possibly making that function more accurate in various numeric situations. As another possible avenue, one might be able to adapt the techniques presented here for use in LPG's heuristic precondition cost estimation process (Gerevini, Serina, Saetti, & Spinoni, 2003b), making it more sensitive to the numeric constructs, and thereby – potentially – further improving LPG's performance.

- Koehler's extension of IPP to a numeric context (Koehler, 1998) suffers from complications in the backward search procedure, which significantly degrade runtime performance. Do the same difficulties arise in the monotonic context?

- Koehler and Hoffmann (2000a) argue that there is a reasonable ordering $B \leq A$ between two goals $A$ and $B$ if, from all states where $A$ is achieved first, one must delete $A$ in order to achieve $B$. Under monotonicity, the straightforward translation of this is that two numeric goals $A$ and $B$ are ordered $B \leq A$ if, once the values of the variables that participate in $A$ are sufficient to achieve $A$, their values must be decreased below the necessary value again in order to achieve $B$. It seems that Koehler and Hoffmann's techniques to approximate STRIPS goal orderings transfer easily to this situation. Similarly, it seems that under monotonicity the definitions and approximation techniques given for landmarks (subgoals that will necessarily arise during planning) by Porteous, Sebastia, and Hoffmann (2001) can directly be transferred to numeric goals.

## Acknowledgments

This article is an extended and revised version of a paper (Hoffmann, 2002a) that has been published at ECAI-02. Metric-FF was developed and implemented while the author was visiting the Durham Planning Group, Durham, UK. Thanks go to Maria Fox and Derek Long for discussions, and for making the stay enjoyable. Thanks also go to Malte Helmert for some fruitful remarks on the relaxation theory. I thank David Smith for his insights into the nature of inverted variables, and for many detailed comments on the language. Finally I thank the anonymous reviewers for their comments, which helped improve the paper.

# Appendix A. Proofs

This appendix presents the proofs to all theorems in detail. There are three different classes of results, which we focus on in turn: relaxation adequacy, relaxed Graphplan completeness, and relaxed Graphplan correctness. Within each of these classes, the results are given for languages of increasing expressivity.

## A.1 Relaxation Adequacy

For STRIPS, the restricted numeric language, and numeric tasks in general, we prove that the respective relaxations are adequate (in the general case, we identify situations where the relaxation is adequate). The proof for the STRIPS case is trivial.

**Proposition 1** *The relaxation given in Definition 1 is adequate, i.e., the following holds true.*

1. *Admissibility: any plan that solves the original task also solves the relaxed task, i.e., assuming a STRIPS task $(P, A, I, G)$, any plan for $(P, A, I, G)$ is also a relaxed plan for $(P, A, I, G)$.*

2. *Basic informedness: the preconditions and goals can trivially be achieved in the original task if and only if the same holds in the relaxed task, i.e., assuming a STRIPS task $(P, A, I, G)$, $\langle \rangle$ is a plan for $(P, A, I, G)$ if and only if $\langle \rangle$ is a relaxed plan for $(P, A, I, G)$, and for $a \in A$, $result(I, \langle \rangle) \supseteq$ pre$(a)$ if and only if $result(I, \langle \rangle) \supseteq$ pre$(a^+)$.*

3. *Polynomiality: the relaxed task can be solved in polynomial time, i.e., deciding RPLANSAT is in P.*

**Proof:** 1. After application of each action in the relaxed action sequence, at least the propositions are true that are true in the real sequence. So each action precondition, and the goal, is fulfilled.

2. Holds because we are not dropping any precondition or goal constraints. The empty plan $\langle \rangle$ is a plan for $(P, A, I, G)$ if and only if $G \subseteq I$ holds. The same is true for $(P, A^+, I, G)$. Similarly for action preconditions.

3. This was proved by Bylander (1994). □

The proof for the case of the restricted numeric language is a straightforward extension to the STRIPS proof, exploiting the correspondence between pre/goal-conditions, add lists, and delete lists on the one hand, and $x \geq [>]c$ constraints, += effects, and -= effects on the other hand.

**Theorem 1** *The relaxation given in Definition 5 is adequate, i.e., the following holds true.*

1. *Admissibility*:  *assuming  a  restricted  numeric  task*  $(V, P, A, I, G)$, *any plan for* $(V, P, A, I, G)$ *is also a relaxed plan for* $(V, P, A, I, G)$.

2. *Basic informedness*:  *assuming a restricted numeric task* $(V, P, A, I, G)$, $\langle\rangle$ *is a plan for* $(V, P, A, I, G)$ *if and only if* $\langle\rangle$ *is a relaxed plan for* $(V, P, A, I, G)$, *and for* $a \in A$, $result(I, \langle\rangle) \models pre(a)$ *if and only if* $result(I, \langle\rangle) \models pre(a^+)$.

3. *Polynomiality*: *deciding RESTRICTED-RPLANSAT is in P.*

**Proof:** 1. After application of each step in the relaxed plan, at least the propositions are true that are true in the real plan, and the values of all numeric variables are at least as high as in the real plan. As all action preconditions and the goal only require variable values to be greater than or equal to a constant, all these constraints remain fulfilled.

2. Holds because we are not dropping any precondition or goal constraints. The empty plan $\langle\rangle$ is a plan for $(V, P, A, I, G)$ if and only if $I \models G$ holds. The same is true for $(V, P, A^+, I, G)$. Similarly for action preconditions.

3. The following is a polynomial time algorithm that decides RESTRICTED-RPLANSAT.

$M := I$, $m := v(I)$
remove, from action preconditions and the goal, all propositions in $M$ and
all numeric constraints that are fulfilled by the $m^i$ values (i.e., $m^i \geq [>]c$)
**while** $G \neq \emptyset$ **do**
$\quad A := \{a \in A \mid pre(a) = \emptyset\}$
$\quad M' := M \cup \bigcup_{a \in A} p(\mathit{eff}(a))^+$
$\quad m' := m$
$\quad$**for** $i \in \{1, \dots, n\}, m^i \neq \infty$ **do**
$\quad\quad$**if** $\exists a \in A : (v^i, +=, c) \in v(\mathit{eff}(a))$ **then** $(m')^i := \infty$ **endif**
$\quad$**endfor**
$\quad$**if** $M' = M$ and $m' = m$ **then** fail **endif**
$\quad M := M'$, $m := m'$
$\quad$remove, from action preconditions and the goal, all propositions in $M$ and
$\quad$all numeric constraints that are fulfilled by the $m^i$ values
**endwhile**
succeed

Remember that $n$ denotes the number of numeric variables. Denote by $A_t$ the action set in iteration $t$ of this algorithm. We prove that the algorithm succeeds if there is a relaxed plan, that there is a relaxed plan if the algorithm succeeds, and that the algorithm takes polynomial time in the size of the task.

If there is a relaxed plan $\langle a_1, \dots, a_k \rangle$ for $(V, P, A, I, G)$, then $a_t \in A_t$ holds true for $1 \leq t \leq k$: the set $M$ (the values $m$) always include (are always at least as high as) the true facts in the relaxed plan (the variable values in the relaxed plan). The algorithm succeeds after at most $k$ iterations. It does not fail earlier as this implies a fixpoint in contradiction to reachability of the goals.

In the other direction, if the algorithm succeeds in an iteration $k$ then one can construct a relaxed plan. Simply linearize the (relaxations of the) actions in the sets $A_1, \dots, A_k$ in an arbitrary order. If an action at a layer $t$ has a $+=$ effect on a variable $x^i$, then repeatedly execute the action until all constraints on $x^i$ that have been removed in iteration $t$ are fulfilled (as all the constraints are of the form $x^i \geq [>]c$, this will eventually happen). The actions applied this way all have their preconditions fulfilled as these were empty at the respective iteration, and the execution sequence makes the same constraints true as the algorithm.

As for runtime, each single iteration is polynomial. An upper bound on the number of iterations is $|V| + |P|$. In each iteration, to avoid failure, at least one new proposition must enter $M$ or one new variable value must be set to $\infty$. □

Generalizing from the restricted language, ignoring the decreasing effects is adequate if all numeric constraints are monotonic, and all numeric effects are strongly monotonic (plus changes due to $:=$ effects can not propagate into a numeric variable's own value). The proof generalizes, in this way, from the proof above.

**Theorem 2** *The relaxation given in Definition 7 is adequate for strongly monotonic tasks with acyclic $:=$ effects, i.e., the following holds true.*

1. Admissibility: *assuming a monotonic numeric task $(V, P, A, I, G)$, any plan for $(V, P, A, I, G)$ is also a relaxed plan for $(V, P, A, I, G)$.*

2. Basic informedness: *assuming a numeric task $(V, P, A, I, G)$, $\langle \rangle$ is a plan for $(V, P, A, I, G)$ if and only if $\langle \rangle$ is a relaxed plan for $(V, P, A, I, G)$, and for $a \in A$ $result(I, \langle \rangle) \models \mathrm{pre}(a)$ if and only if $result^+(I, \langle \rangle) \models \mathrm{pre}(a)$.*

3. Polynomiality: *deciding STRONGLY-MONOTONIC-RPLANSAT is in P.*

**Proof:** 1. Say $\langle a_1, \ldots, a_n \rangle$ is a plan for $(V, P, A, I, G)$. Executing the sequence under $result$, all precondition and goal constraints are fulfilled. Denote by $v^i(t)$ the value of variable $i$ after execution of action $a_t$, and denote by $v^i(t)^+$ the value of variable $i$ after execution of action $a_t$ under $result^+$. We show that $v^i(t) \leq v^i(t)^+$ for all $i$ and $t$. With monotonicity of numeric constraints, Definition 8 condition (1), this suffices. The claim is easily shown by induction over $t$. With $t = 1$, $v^i(1) \leq v^i(1)^+$ holds for all $i$ simply because $result^+$ is identical to $result$ except that all effects that decrease the value of a variable are ignored. From $t$ to $t + 1$, if $v^i(t) \leq v^i(t)^+$ for all $i$ then $v^i(t + 1) \leq v^i(t + 1)^+$ holds for all $i$ due to the same argument, plus the monotonicity of the numeric effects in the sense of Definition 8 condition (2): the higher the input numeric variables are, the higher the resulting value of the affected variable becomes.

2. The empty plan $\langle \rangle$ is a plan for $(V, P, A, I, G)$ if and only if $I \models G$ holds. The same is true for $\langle \rangle$ as a relaxed plan, as we are not dropping any goal constraints. Similarly for action preconditions.

3. The following is a polynomial time algorithm that decides relaxed solvability of a strongly monotonic task with acyclic $:=$ effects.

1. $M := I$, $m := v(I)$
2. remove, from action preconditions and the goal, all propositions in $M$ and

all numeric constraints that are fulfilled by the $m^i$ values

3.   **while** $G \neq \emptyset$ **do**

4.      $A := \{a \in A \mid pre(a) = \emptyset\}$

5.      $M' := M \cup \bigcup_{a \in A} p(\mathit{eff}(a))^+$

6.      $m' := m$

7.      **for** $i \in \{1, \ldots, n\}, m^i \neq \infty$ **do**

8.         **if** $\exists a \in A, (v^i, ass, exp) \in v(\mathit{eff}(a))$ :

9.            $ass \in \{+=, -=, *=, /=\}, (v^i, ass, exp)(m) > m^i$ **then** $(m')^i := \infty$ **endif**

10.     **endfor**

11.     **for** $i \in \{1, \ldots, n\}, m^i \neq \infty$ **do**

12.        **if** $\exists a \in A, (v^i, :=, exp) \in v(\mathit{eff}(a)) : (v^i, :=, exp)(m) > m^i$ **then**

13.           $(m')^i := \max_{a \in A, (v^i, :=, exp) \in v(\mathit{eff}(a)):(v^i, :=, exp)(m) > m^i} (v^i, :=, exp)(m)$

14.        **endif**

15.     **endfor**

16.     **if** $M' = M$ and $m' = m$ **then** fail **endif**

17.     $M := M', m := m'$

18.     remove, from action preconditions and the goal, all propositions in $M$ and all numeric constraints that are fulfilled by the $m^i$ values

19.  **endwhile**

20.  succeed

Here, as above, $v(exp)$ for an expression $exp$ denotes the set of all variables contained in $exp$. The value of an expression that contains variables set to infinity is given as the limit of the expression in these variables. Note that by assumption the limits are all $\infty$ (Definition 8 condition (4)) so they can, in particular, be computed efficiently. We prove that the algorithm succeeds if there is a relaxed plan, that there is a relaxed plan if the algorithm succeeds, and that the algorithm takes polynomial time in the size of the task.

Denote by $A_t$ the action set in iteration $t$ of the algorithm. If there is a relaxed plan $\langle a_1, \ldots, a_k \rangle$ for $(V, P, A, I, G)$, then $a_t \in A_t$ holds true for $1 \leq t \leq k$: the variable updates on $m$ performed in the algorithm are always at least as high as those performed by the $result^+$ function. Note here that line 13 takes the maximum over the available $:=$ effects. Note also that all effects obey Definition 8 condition (2), so one needs consider only the maximum input values in order to obtain the maximum output value. In consequence, with monotonicity of numeric constraints in the sense of Definition 8 condition (1), the algorithm reaches the goals and succeeds after at most $k$ iterations. It does not fail earlier as this implies a fixpoint in contradiction to reachability of the goals.

If the algorithm succeeds after an iteration $k$ then one can construct a relaxed plan as follows. Perform an upwards loop from 1 to $k$. At each iteration $t$, repeatedly apply all actions in $A_t$ until all the constraints that have been removed in line 18, in iteration $t$, are fulfilled. We show below that this point will eventually be reached. Once the point is reached, one can continue with the next higher $t$ value until the step at the succeeding iteration $k$ has been completed. All the actions applied this way have their preconditions fulfilled as these were all empty at the iteration $t$ where the actions are applied, as the constructed relaxed plan always fulfills the same constraints that were removed in an iteration, and as by Definition 8 condition (1) constraints can not become false again once they are true in a relaxed plan. For the same reason the goals are fulfilled at the end of iteration $k$. It remains to show that, at an iteration $t$, repeatedly applying the actions in $A_t$ will eventually fulfill all the constraints removed in that iteration. Denote by $I_t$ the set of all variables that got set, in iteration $t$, to $\infty$ in line 9, denote by $I'_t$ the set of all variables that got set to $\infty$ in line 13, and denote by $F_t$ the set of all variables that got set to a new value below $\infty$ in line 13. We show that:

1. After one application of the actions in $A_t$ the variables in $F_t$ have at least the values that they had at constraint removal in line 18.

2. With repeated application of the actions in $A_t$ the variables in $I_t \cup I'_t$ reach arbitrarily high values.

This suffices for the constraints eventually being fulfilled. Assume the two claims hold true. Then, with monotonicity of the constraints (Definition 8 condition (1)) the variables in $F_t$ contribute at least as much to the fullfillment of these constraints as they did in iteration $t$ of the decision algorithm. As for the variables in $I_t \cup I'_t$, there is a finite assignment to these variables, higher than their previous values, that makes the respective constraints true at this point. This is a simple consequence of Definition 8 condition (1) (the constraints prefer higher variable values), condition (4) (the expressions diverge in the variables), condition (5) (existence of a finite fulfilling assignment), and the fact that the constraints were not true in the previous iteration but became true when setting the variables in $I_t \cup I'_t$ to $\infty$.

The first claim follows from the simple fact that the actions responsible for increasing the values of the variables in $F_t$ – the actions that fulfill the condition in line 12 – are, in particular, contained in $A_t$. Their outcome might be higher if other variables in the respective effect right hand side have been increased first; there are no negative interactions with other variables as we are considering the relaxed transition function. The argument for the second claim is as follows. As for the variables in $I_t$, $A_t$ contains the respective responsible action fulfilling the condition in lines 8 and 9. Each application of this action increases, by Definition 8 condition (3), the variable's value by at least as much as the previous application, so repeated application diverges. Note that, again, under relaxed state transition, applying an action can not worsen the situation for other variables. As for the variables in $I'_t$, $A_t$ contains the action $a$ fulfilling the condition in line 13, with $(v^i, := , exp) \in v(\mathit{eff}(a))$, $exp$ containing at least one variable $v' \in v(exp)$ set to $\infty$ at this point (as $(v^i, := , exp)(m) = \infty$). Recursively, a responsible action $a'$ setting $v'$ to $\infty$ must have been included in the previous iteration. If the effect of $a'$ on $v'$ is a $:=$ effect, a responsible action must have been included earlier, and so on. At one point, the responsible action $a''$ for the respective ancestor variable $v''$ must have been included in line 9. Repeated application of $a''$ causes the value of $v''$ to diverge (with the same argument as above), and in effect transitively causes the value of $v^i$ to diverge.

It finally remains to show that the algorithm terminates in polynomial time. Obviously each single iteration is polynomial. The number of iterations is bounded by the number of times that $M'$ or $m'$ can be different from $M$ respectively $m$. Changes to these values occur in lines 5, 9, and 13. The overall number of changes in line 5 is bound by the number of logical propositions, $|P|$. The overall number of changes in line 9 is bound by the number of numeric variables, $|V|$. So if there was an exponential number of iterations until termination then there would be an exponential number of consecutive iterations where changes occur only in line 13. The number of such consecutive iterations is, however, bound by $|V| * |A|$. This can be seen as follows. Throughout the entire sequence of iterations, only := effects contribute to the changes. The := effects are acyclic by our assumption so their value change can not propagate into their own value, and the only possible further change can occur when a new action comes in. It takes at most $|V|$ iterations to propagate the changes through all variables (this is the length of the longest possible propagation path), so, if at least one new action comes in at an iteration $t$, then another new action comes in at iteration $t + |V|$ at the latest. The obvious bound on the number of iterations where new actions come in is $|A|$, which concludes the argument. □

## A.2 Relaxed Graphplan Completeness

For STRIPS and LNF tasks we prove that the respective relaxed Graphplan mechanisms are complete, i.e., that they find a relaxed plan if there is one. The proof for the STRIPS case is trivial.

**Proposition 2** *Assume a STRIPS task $(P, A, I, G)$, and a state $s$. If the algorithm depicted in Figure 1 fails, then there is no relaxed plan for $(P, A, s, G)$.*

**Proof:** We show the contrapositive, i.e., if there is a relaxed plan for $(P, A, s, G)$, then the algorithm succeeds. Say there is a relaxed plan $P = \langle a_1, \ldots, a_m \rangle$ for $(P, A, s, G)$. The algorithm applies, at the first layer, all possible actions. In particular, this includes $a_1$, so at layer $P_1$ at least the facts are true that are true after executing the first step in $P$. The same argument can inductively be applied for all actions in $P$, implying that at each layer $t$ we have $a_t \in A_t$, and $P_t$ contains all facts that are true upon execution of the first $t$ actions in $P$. This implies that the goals are true at some layer $m' \leq m$, $G \subseteq P_{m'}$. Moreover, the algorithm does not fail at any layer $m'' < m'$: if so then it follows that a fixpoint is reached, $P_i = P_{m''}$ for all $i > m''$, so $G \nsubseteq P_{m'}$, which contradicts our assumptions. □

The proof for LNF tasks proceeds along the same line, but requires some care with the details concerning the values beyond which numeric variables can no longer contribute to a solution.

**Theorem 3** *Assume a linear numeric task $(V, P, A, I, G)$ that is in LNF and has acyclic := effects. Assume a state $s$. If the algorithm depicted in Figure 6 fails, then there is no relaxed plan for $(V, P, A, s, G)$.*

**Proof:** We show the contrapositive, i.e., if there is a relaxed plan for $(V, P, A, s, G)$, then the algorithm succeeds. Say there is a relaxed plan $P = \langle a_1, \ldots, a_m \rangle$ for $(V, P, A, s, G)$. The algorithm applies, at the first layer, all possible actions. In particular, this includes $a_1$, so at layer $P_1$ at least the facts are true that are true after executing the first step in $P$, and the $max_1^i$ values are at least as high as the respective variable values. Together with the fact that the effect right hand sides are positively monotonic (so inserting the $max_t$ values can only increase the outcome), the same argument can inductively be applied for all actions in $P$, implying that at each layer $t$ we have $a_t \in A_t$, $P_t$ contains all facts that are true upon execution of the first $t$ actions in $P$, and the $max_t^i$ values are at least as high as the respective variable values. This, with the monotonicity of the numeric constraints, implies that the goals will be reached at some layer $m' \leq m$, $p(G) \subseteq P_{m'}$ and for all $(exp, \geq [>], 0) \in v(G) : exp(max_{m'}) \geq [>]0$. Moreover, the algorithm does not fail at any layer $m'' < m'$. Assume it does. Then at $m''$ no new propositions have come in, and the $max^i$ values all either have not changed, or are already above their maximum needed value. Denote by $L$ the set of variables $v^i$ whose value is still too low, $max^i_{m''} \leq \text{mneed}^i(s)$. Note that $L \subseteq rV$ holds since outside $rV$ the mneed values are $-\infty$. We have $P_{m''+1} = P_{m''}$ and, for all $v^i \in L$, $max^i_{m''+1} = max^i_{m''}$. We show that $P_{m''+2} = P_{m''+1}$ and, for all $v^i \in L$, $max^i_{m''+2} = max^i_{m''+1}$. This proves the claim: by iterating the argument, the same holds true at all layers $t > m'' + 1$, and we get a contradiction to the goals being reached at $m'$ (note that all constraints in which a variable out of $V \setminus L$ participates are already fulfilled, so increasing these variables can not reach new goal constraints). The set of propositions could increase at layer $m'' + 2$ if a new action came in, i.e., if there was $a \in A_{m''+1}$, $a \notin A_{m''}$. The value of a variable $v^i \in L$ could increase at layer $m'' + 2$ if: a new action came in; a $+=$ effect right hand side expression $(v^i, +=, exp)$ became positive in $A_{m''+1}$ as a result of increasing the $V \setminus L$ variable values from $m''$ to $m'' + 1$; a $:=$ right hand side expression $(v^i, :=, exp)$ in $A_{m''+1}$ became higher than $max^i_{m''+1}$ as a result of increasing the $V \setminus L$ variable values from $m''$ to $m'' + 1$. None of these three cases can occur by definition of the mneed values (that the variables in $V \setminus L$ have reached). As for the first case, $A_{m''+1}$ can not contain a new action because no new precondition constraints became true from $m''$ to $m'' + 1$ – only the $V \setminus L$ variable values have increased, and the constraints in which these participate are already fulfilled at $m''$. As for the second case, all $(v^i, +=, exp)$ effect right hand sides in which $V \setminus L$ variables participate are already above 0 with the values at $m''$ ($v^i \in L \subseteq rV$, so the mneed definition for $+=$ effects applies). As for the third case, if this occurred then there was at least one variable $v^j \in V \setminus L$ contained in the right hand side of the responsible effect $(v^i, :=, exp)$. This variable would fulfill $max^j_{m''} > \text{mneed}^j(s)$, thus $exp(max_{m''}) > \text{mneed}^i(s)$ would hold ($v^i \in L \subseteq rV$, so the mneed definition for $:=$ effects applies), thus $max^i_{m''+1} > \text{mneed}^i(s)$ would hold (through application of $(v^i, :=, exp)$ in $A_{m''}$) in contradiction to our assumptions. This concludes the argument. $\qquad\square$

## A.3 Relaxed Graphplan Correctness

For STRIPS and LNF tasks we prove that the respective relaxed Graphplan mechanisms are correct, i.e., that the actions they select form a relaxed plan. The proof for the STRIPS case is trivial.

**Proposition 3** *Assume a STRIPS task $(P, A, I, G)$, and a state $s$ for which the algorithm depicted in Figure 1 reaches the goals. The actions selected by the algorithm depicted in Figure 2 form a relaxed plan for $(P, A, s, G)$.*

**Proof:** First, note that at each layer $t$ and for each goal $g \in G_t$, there is at least one action $a$ such that $level(a) = t - 1$, $g \in \text{eff}(a)^+$, due to the way the levels are computed. Also, an action's preconditions always have a lower level than the action itself.

The algorithm selects a set $A_t$ at each layer $t$. We can arrange the actions in each of these sets in an arbitrary order to obtain a relaxed plan for $(P, A^+, s, G)$. All goals and sub-goals at a layer $t$ are achieved by the actions in $A_{t-1}$. So with delete effects being ignored, at least the propositions are true which are needed. □

The proof for LNF tasks is a straightforward extension of the STRIPS proof.

**Theorem 4** *Assume a linear numeric task $(V, P, A, I, G)$ that is in LNF and has acyclic := effects. Assume a state $s$ for which the algorithm depicted in Figure 6 reaches the goals. The actions selected by the algorithm depicted in Figure 7 form a relaxed plan for $(V, P, A, s, G)$.*

**Proof:** First, note that at each layer $t$ and for each goal $g \in G_t$, there is at least one action $a$ such that $level(a) = t - 1$, $g \in \textit{eff}(a)^+$, due to the way the levels are computed. For the numeric goals $(exp, \geq [>], 0) \in v(G_t)$, there is always a := effect with sufficiently high right hand side value, or

$$max_t^i - \sum_{a \in A_t : (v^i, +=, exp) \in v(\textit{eff}(a)), exp(max_{t-1}) > 0} exp(max_{t-1}) = max_{t-1}^i$$

holds. In the first case the **while** loop is not entered, in the second case it terminates successfully. Note that one occurrence of an action can support different logical and numeric goals by different effects, but can not be used to support the same numeric goal twice.

Denote, for a layer $t$, by $A_t$ the set of actions selected by the algorithm at that layer. We can arrange the actions in each of these sets in an arbitrary order to obtain a relaxed plan for $(A^+, s, G)$. All goals and sub-goals at a layer $t$, both logical and numeric, are achieved by the actions in $A_{t-1}$. The expressions in numeric goals and the effect right hand sides are always at least as high as required as we constrain all contained variables to take on their respective maximum values. With delete effects being ignored, at least the propositions are true which are needed. With decreasing effects being ignored and monotonicity of effect right hand sides, the expression values in constraints are at least as high as required. □